\documentclass[10pt,twocolumn,letterpaper]{article}

\usepackage[pagenumbers]{cvpr} 

\usepackage{booktabs}
\usepackage{balance}
\definecolor{cvprblue}{rgb}{0.21,0.49,0.74}
\usepackage[pagebackref,breaklinks,colorlinks,allcolors=cvprblue]{hyperref}
\usepackage{wrapfig}
\usepackage{booktabs}
\usepackage{hyperref}
\usepackage{url}
\usepackage{lipsum} % for placeholder text
\usepackage{float}
\usepackage{graphicx}
\usepackage{subcaption}
\usepackage{multirow}

\usepackage[table]{xcolor} 
\usepackage{xfp}
\usepackage{pgf,tikz} 
\usepackage{amsmath}
\usepackage{enumitem}
\usepackage{makecell}
\usepackage{tabularx}
\usepackage{placeins}
\usepackage{caption}
\usepackage[export]{adjustbox}
\usepackage{subcaption}
\usepackage{csquotes}
\usepackage{titletoc}
\newcommand{\paragrapht}[1]{\noindent\textbf{#1}}
\usepackage{dblfloatfix} 
\usepackage{graphicx}

\newcommand\blfootnote[1]{%
  \begingroup
  \renewcommand\thefootnote{}\footnote{#1}%
  \addtocounter{footnote}{-1}%
  \endgroup
}

\def\paperID{2178} % *** Enter the Paper ID here
\def\confName{CVPR}
\def\confYear{2026}

\title{\raisebox{-0.25\height}{\includegraphics[height=2em]{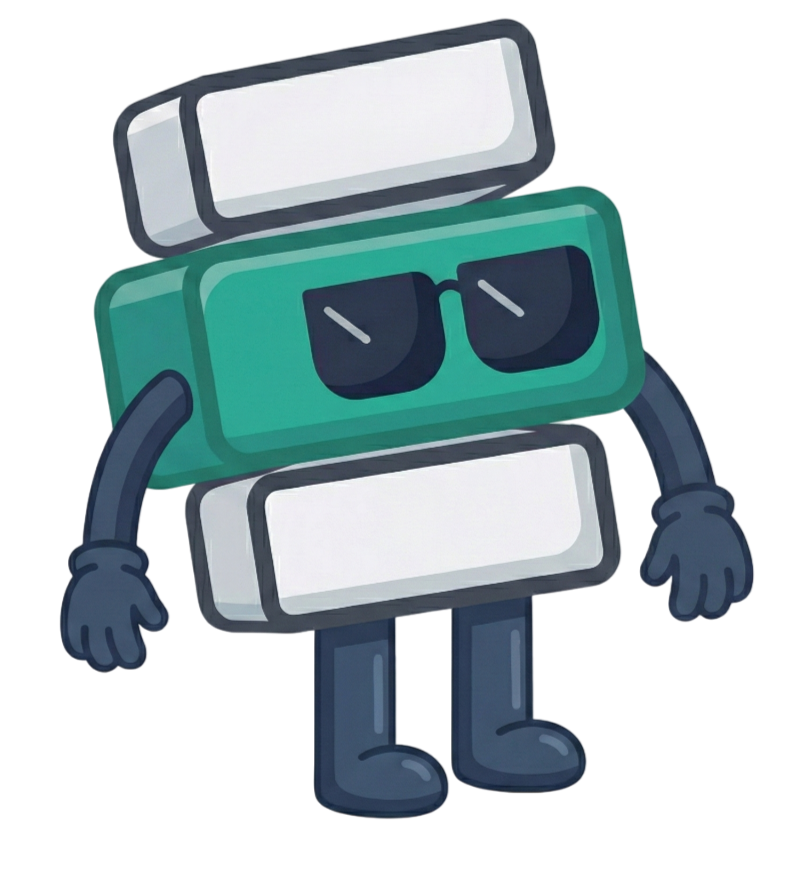}}
Correspondence-Attention Alignment for Multi-View Diffusion Models}

\author{
Minkyung Kwon$^{*1}$ \quad
Jinhyeok Choi$^{*1}$ \quad
Jiho Park$^{*1}$ \\
Seonghu Jeon$^{1}$ \quad
Jinhyuk Jang$^{1}$ \quad
Junyoung Seo$^{1}$ \quad
Minseop Kwak$^{1}$ \quad \\
Jin-Hwa Kim$^{\dag2,3}$ \quad 
Seungryong Kim$^{\dag1}$ \\ \\ 
$^{1}$KAIST AI\quad$^{2}$NAVER AI Lab\quad$^{3}$SNU AIIS\\
}

\begin{document}
\twocolumn[{%
\renewcommand\twocolumn[1][]{#1}%
\maketitle
\begin{center}
    \centering
    \vspace{-20px}
    \captionsetup{type=figure}
    \includegraphics[width=1\textwidth]{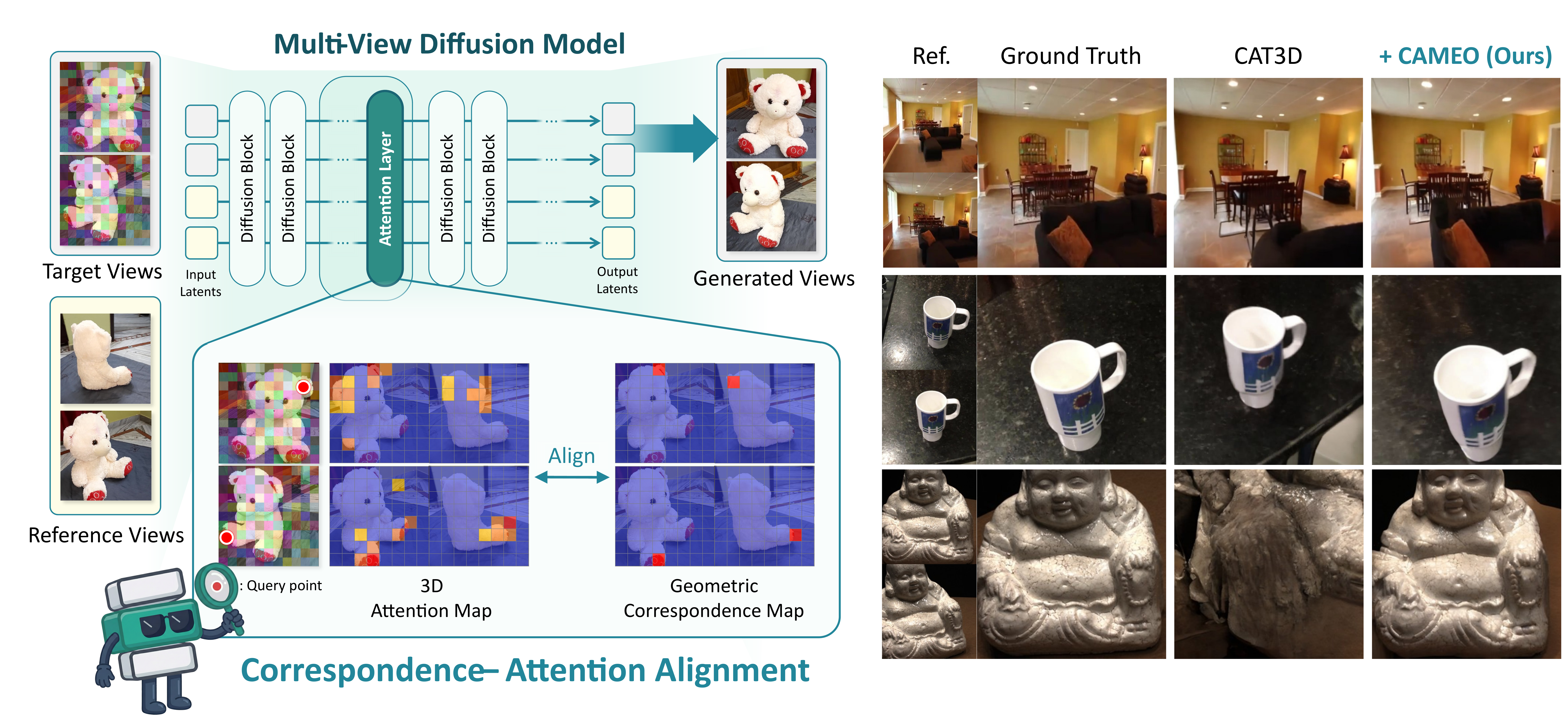}
    \vspace{-20pt}
    \captionof{figure}{\textbf{Correspondence-attention alignment makes multi-view diffusion training effective.} Our framework, \textbf{CAMEO}, aligns attention maps of the multi-view diffusion models~\cite{cat3d,mvgenmaster,li2024hunyuandit} with geometric correspondence. In experiments, {CAMEO} produces geometrically consistent novel views even in challenging scenarios involving large viewpoint changes or complex geometry. }
    \label{fig:teaser}
\end{center}
}]

\blfootnote{*: Equal contribution}
\blfootnote{\dag: Co-corresponding author}

% \begin{document}
% \maketitle
\begin{abstract}
Multi-view diffusion models have recently emerged as a powerful paradigm for novel view synthesis, yet the underlying mechanism that enables their view-consistency remains unclear. In this work, we first verify that the attention maps of these models acquire geometric correspondence throughout training, attending to the geometrically corresponding regions across reference and target views for view-consistent generation. However, this correspondence signal remains incomplete, with its accuracy degrading under large viewpoint changes. Building on these findings, we introduce \textbf{CAMEO}, a simple yet effective training technique that directly supervises attention maps using geometric correspondence to enhance both the training efficiency and generation quality of multi-view diffusion models. Notably, supervising a single attention layer is sufficient to guide the model toward learning precise correspondences, thereby preserving the geometry and structure of reference images, accelerating convergence, and improving novel view synthesis performance. {CAMEO} reduces the number of training iterations required for convergence by half while achieving superior performance at the same iteration counts. We further demonstrate that {CAMEO} is model-agnostic and can be applied to any multi-view diffusion model. Project page is available at \url{https://cvlab-kaist.github.io/CAMEO/}.
\end{abstract}    
\section{Introduction}
\label{sec:intro}

Novel view synthesis (NVS) is the task of predicting images from unseen viewpoints given reference views, preserving geometric consistency and photorealistic appearance. While optimization-based methods~\cite{3dgs,nerf} rely on per-scene optimization with dozens of input images, recent generative approaches, including multi-view diffusion models~\cite{3dim, zero123, zero123++, cat3d, mvgenmaster, bolt3d, stablevirtualcamera, matrix3d}, leverage generative priors from large-scale 2D diffusion models~\cite{ho2020denoising, ldm} to synthesize novel views. 

Multi-view diffusion models~\cite{shi2023mvdream,cat3d,mvgenmaster} employ 3D self-attention to aggregate information across viewpoints, where each query token from one view attends to all spatial locations across all views. As a result, they generate novel-view images that exhibit geometric consistency across views. However, as illustrated in \cref{fig:teaser}, the consistency often deteriorates in challenging scenarios involving large viewpoint changes or complex geometry, leading to cross-view misalignment and structural degradation. These observations motivate a central question: \textit{how do multi-view diffusion models maintain view-consistency through their internal mechanisms?} 

In this work, we aim to understand the internal mechanisms governing geometric consistency in multi-view diffusion models~\cite{shi2023mvdream,cat3d,mvgenmaster} and leverage this understanding to improve their performance. To achieve this, we first analyze their 3D self-attention maps, yielding three key findings that motivate our method. (1) These models learn to encode geometric correspondences within their attention layers, and this emergent property concentrates in particular layers of the model. (2) The quality of this learned correspondence directly impacts the models' performance: correspondence improves throughout training and correlates strongly with generation quality. (3) Yet, this correspondence signal is limited.  A substantial precision gap remains compared to geometry prediction models (\eg VGGT~\cite{wang2025vggt}), and the correspondence fails under large viewpoint rotations. These observations suggest that \textit{while geometric correspondence is inherently learned within the attention layers of multi-view diffusion models, the signal remains incomplete and fragile.}

Our findings point to a promising direction: augment training with explicit geometric supervision to mitigate these limitations and enhance the synthesis quality.
To this end, we introduce \textbf{CAMEO} (\textbf{C}orrespondence–\textbf{A}ttention Alignment for \textbf{M}ulti-vi\textbf{e}w Diffusi\textbf{o}n Models), a simple yet effective technique that supervises attention layers with geometric correspondence. We demonstrate that supervising a single attention layer is sufficient to improve both learning efficiency and performance.

CAMEO provides strong cues for geometrically consistent generation, resulting in faster convergence and higher-quality NVS. Notably, our method preserves geometry—generated images maintain accurate shapes consistent with the reference views, as illustrated in~\cref{fig:teaser}.

% Paragraph 6 (Ver 1.)
To evaluate the effectiveness, we conduct comprehensive experiments primarily based on CAT3D~\cite{cat3d}, evaluating on both scene-level~\cite{re10k} and object-centric~\cite{reizenstein21co3d,jensen2014largedtu} datasets.  CAMEO reduces the training iterations required for convergence by half, while achieving better performance at the same number of training iterations. To demonstrate its model-agnostic applicability, we further apply CAMEO to a state-of-the-art model~\cite{mvgenmaster}, which utilizes geometric conditions, and a DiT-based~\cite{li2024hunyuandit} multi-view diffusion model, achieving consistent improvements across all frameworks.

The main contributions of this paper are as follows:
\begin{itemize}[itemsep=0.5ex, parsep=0pt, topsep=0.5ex,]
    \item We present an in-depth analysis of multi-view diffusion models, revealing that geometric correspondence emerges in attention maps during training and is critical for NVS performance.
    \item We propose CAMEO, a simple and effective technique that supervises the model's attention maps with geometric correspondence.
    \item We demonstrate that our method enhances NVS performance across representative multi-view diffusion models~\cite{cat3d,mvgenmaster,li2024hunyuandit}, preserving geometry, accelerating convergence, and improving novel view synthesis performance.
\end{itemize}

\section{Related work}
\label{sec:related_work}
\begin{figure*}[t]
  \centering
  \vspace{-6pt}
  % --- Left: 3D attention map ---
  \begin{subfigure}[t]{0.35\textwidth}
    \centering
    \raisebox{0.02\height}{
    \includegraphics[width=\linewidth]{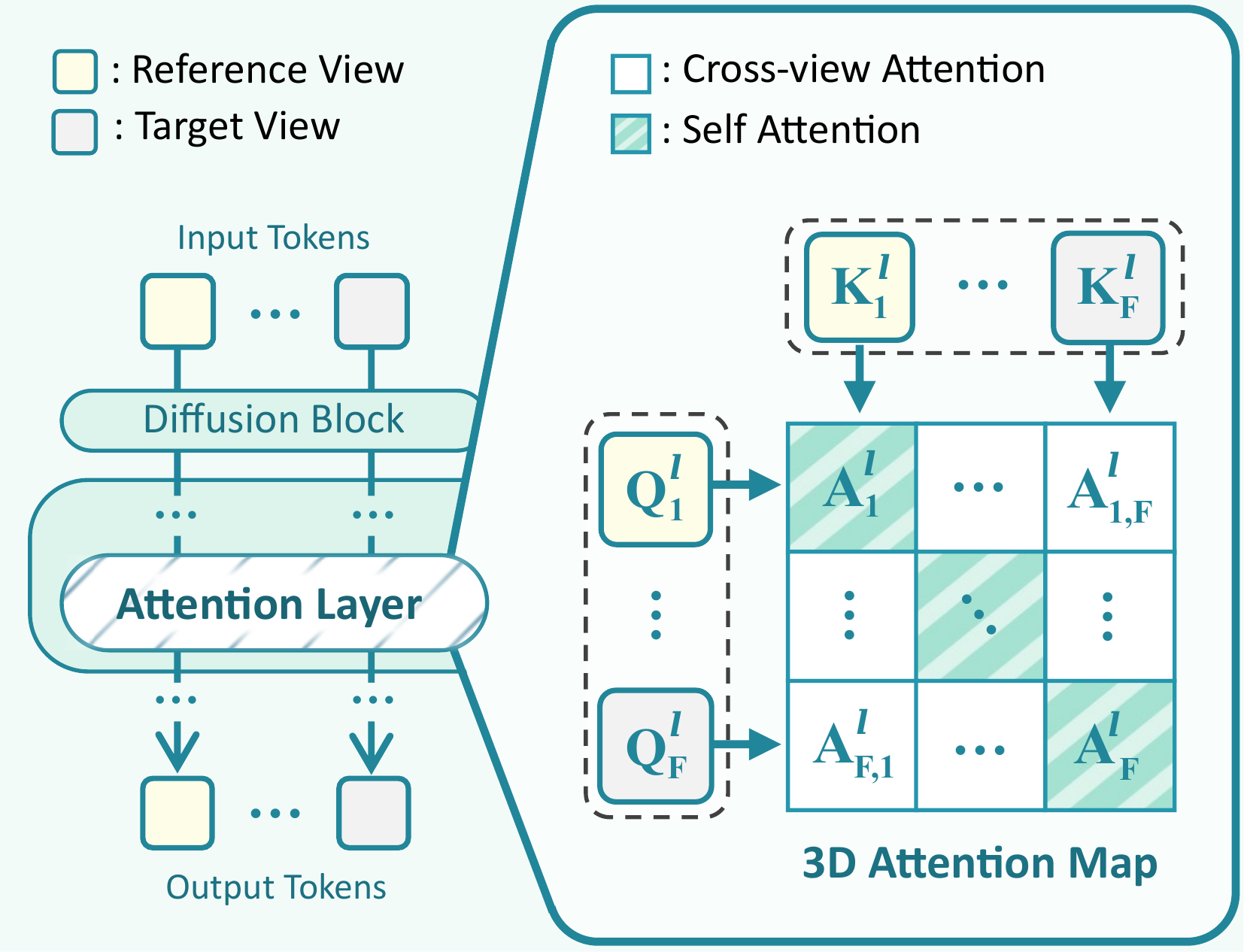}}
    \caption{{3D attention map}}
    \label{fig:3d_attn_map}
  \end{subfigure}
  \hfill
  % --- Right: visualization ---
  \begin{subfigure}[t]{0.63\textwidth}
    \centering
    \includegraphics[width=\linewidth]{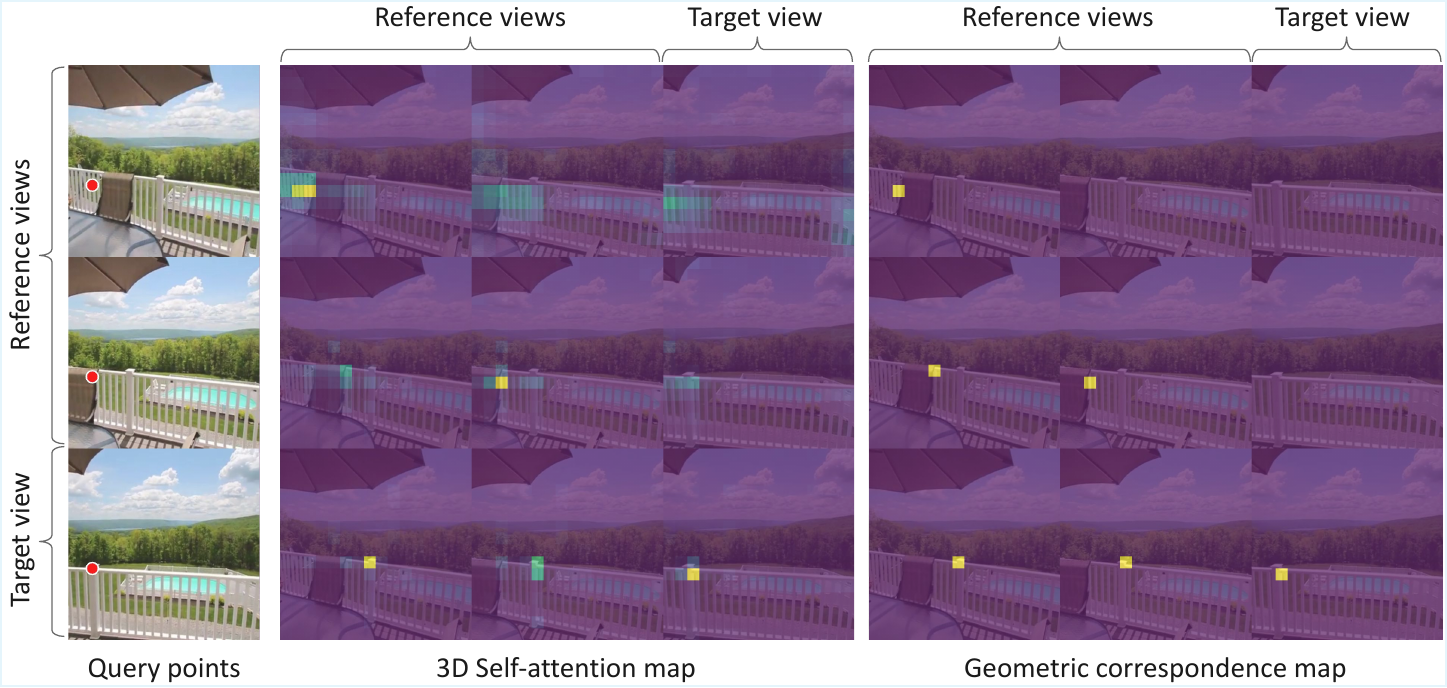}
\caption{{Attention map and correspondence map visualization}}
    \label{fig:attention_and_corr_map}
  \end{subfigure}
  \vspace{-6pt}
  \caption{\textbf{Attention maps in multi-view diffusion models and geometric correspondence map:}
(a) Multi-view diffusion models and their 3D self-attention maps~\cite{cat3d,mvgenmaster}. (b) Attention vs. geometric correspondence map. The attention map of layer $l=10$ in CAT3D~\cite{cat3d} naturally focuses on its geometric counterpart across views even without explicit supervision.}
  \label{fig:3d_attn_and_vis}
  \vspace{-15pt}
\end{figure*}

\paragrapht{Diffusion models for novel view synthesis.}
Diffusion models~\cite{ho2020denoising, ldm} have recently emerged as powerful generative priors for novel view synthesis (NVS), advancing beyond traditional geometry-based approaches~\cite{nerf,3dgs}.
Early methods~\cite{3dim, zero123, zero123++} formulated NVS as a conditional image-to-image translation problem, predicting a single target view from one or more reference views.
More recent multi-view diffusion frameworks synthesize sets of geometrically consistent views by extending 2D latent diffusion models with a 3D self-attention~\cite{cat3d, bolt3d, stablevirtualcamera}. Building upon these, some methods further incorporate conditioning from geometry-prediction models~\cite{moai, trajectorycrafter, matrix3d}.

We focus on analyzing the underlying mechanisms by which diffusion models internalize geometric cues, showing that correspondence in attention enables effective inference of geometry. We find that explicitly aligning attention with geometric correspondence further improves view-consistent generation.

\paragrapht{Learning feature representations by supervision.}
Recent work improves diffusion models~\cite{ldm,dit,Blattmann2023StableVD} by supervising their internal features with external signals. REPA~\cite{repa} introduces feature-level supervision by distilling semantic features from DINOv2~\cite{oquab2023dinov2} into early layers of a Diffusion Transformer (DiT)~\cite{dit}, accelerating convergence and enhancing semantic structure. Similarly, NVComposer~\cite{li2025nvcomposer} supervises the features using pointmaps obtained from a dense stereo model~\cite{wang2024dust3r}, and Track4Gen~\cite{track4gen} applies tracking supervision on the features to reduce appearance drift. Concurrently, Geometry Forcing~\cite{wu2025geometry} aligns diffusion features with geometry‐aware embeddings from a pretrained geometry prediction model~\cite{wang2025vggt}. These methods show that structural or temporal priors can be encoded through feature alignment, yet the mechanism by which aligned features improve generation remains unclear.

Feature alignment enriches per-view semantics or geometry but does not enforce cross-view consistency~\cite{crepa}. In this work, we identify correspondence in attention as the key signal for consistency. By aligning attention maps rather than features~\cite{repa,wu2025geometry}, we guide the model to attend to geometrically corresponding regions across views, enabling it to capture pose and spatial relationships that improve both consistency and generation quality.

\paragrapht{Attention mechanism in diffusion models.}
Diffusion models commonly employ U-Net~\cite{ldm} and Transformer~\cite{dit} architectures, whose attention mechanisms selectively integrate information across modalities, spatial locations, and temporal sequences, demonstrating an inherent understanding of spatial structure and geometry. In text-to-image diffusion models, attention maps link text prompts to specific spatial regions, allowing localized edits and geometric control~\cite{hertz2022prompt,brooks2023instructpix2pix,nichol2021glide,chefer2023attendandexcite}. This spatial awareness extends to segmentation tasks, where attention accurately separates spatial components~\cite{sun2024iseg}. For video generation, temporal attention naturally encodes tracking information across frames~\cite{nam2025emergenttemporalcorrespondencesvideo}, enabling the generation of motion-realistic videos by capturing object and scene geometry. These observations reveal that attention carries structured spatial and temporal signals. Extending this principle to multi-view settings, we study attention in multi-view diffusion models as a carrier of geometric correspondence across views and introduce explicit supervision to reinforce this correspondence for improved view consistency.
\section{Method}
In the following sections, we first provide preliminaries on multi-view diffusion models (\cref{sec:preliminaries}). We then analyze the attention maps of multi-view diffusion models (\cref{sec:analysis}). Motivated by the findings in our analysis, we propose \textbf{CAMEO}, which accelerates the learning of accurate cross-view relationships and improves novel view synthesis quality (\cref{sec:method}).

\subsection{Preliminaries}
\label{sec:preliminaries}
\begin{figure*}
    \includegraphics[width=\linewidth]{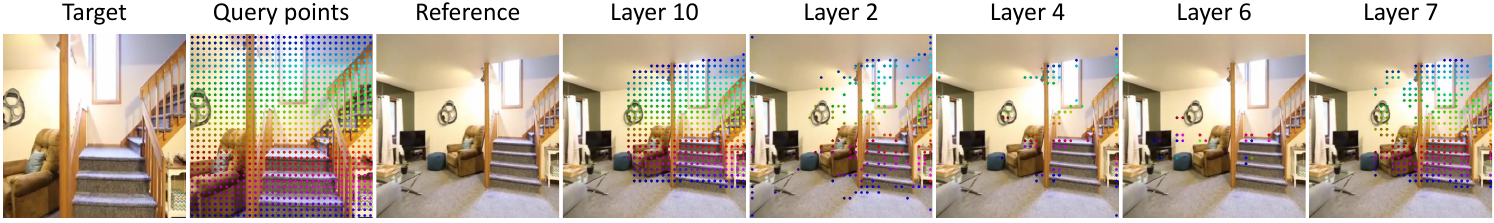}
    \vspace{-15pt}
    \caption{\textbf{Layer-wise behavior of the multi-view diffusion model (CAT3D~\cite{cat3d})'s attention map.} For each query point on the target image, model's maximum attending point in the reference image is marked with the same color as the query point. Attention map of layer $l=10$ cleary attends to geometrically corresponding point, while other layers do not. We fix a timestep $t=999$ (\ie, complete noise).}
    \label{fig:layerwise}
    \vspace{-15pt}
\end{figure*}

The goal of novel view synthesis (NVS) is to generate \(M\) target images
\(\{\mathbf{I}_i^{\text{tgt}}\}_{i=1}^M\) for target
camera poses \(\{\boldsymbol{\pi}_i^{\text{tgt}}\}_{i=1}^M\), given \(N\) reference images
\(\{ \mathbf{I}_i^{\text{ref}}\}_{i=1}^N\) and their
corresponding camera poses \(\{\boldsymbol{\pi}_i^{\text{ref}}\}_{i=1}^N\).

Multi-view diffusion models~\cite{cat3d,mvgenmaster,shi2023mvdream,stablevirtualcamera} built upon pretrained text-to-image diffusion models~\cite{ldm,li2024hunyuandit} typically employ a 3D self-attention mechanism. Specifically, they extend the standard 2D attention in pretrained text-to-image diffusion models by concatenating the token sequences from each view, allowing features to interact both within and across views. Given \(N\) reference views and \(M\) target views, we set the number of total views to \(F = N + M\).
At diffusion timestep \(t\) and attention layer \(l\) of a multi-view diffusion model, features from each of the \(F\) images are projected into query \((\mathbf{Q}_i^{l,t})\) and key \((\mathbf{K}_i^{l,t})\) matrices
for view \(i \in \{1,\ldots,F\}\), each of size \(\mathbb{R}^{hw \times d}\), where $h$ and $w$ denote the height and width of the feature map, and $d$ is the embedding dimension of each token. These are then concatenated along the spatial axis, stacking tokens from all $F$ views into a single sequence, with $N$ reference views followed by $M$ target views. This produces the final query and key matrices, $\mathbf{Q}^{l,t}$ and $\mathbf{K}^{l,t}$, each of size $ \mathbb{R}^{Fhw\times d}$. The 3D attention map \(\mathbf{A}^{l,t}\in \mathbb{R}^{Fhw \times Fhw}\) is then computed via scaled dot-product attention with row-wise softmax, $\texttt{softmax}(\cdot)$. For notational simplicity, we omit the timestep notation and write \(\mathbf{A}^{l}\) henceforth.

As shown in~\cref{fig:3d_attn_map}, the 3D self-attention map can be categorized into two key interactions:
(1) self-attention \(\mathbf{A}_{i}^{l}\), (2) cross-view attention \(\mathbf{A}_{i,j}^l\), where \(i, j \in \{1,\ldots, F\}\) with \(i\not=j\). Of these, cross-view attention \(\mathbf{A}_{i,j}^{l} \in \mathbb{R}^{hw\times{hw}}\) is of particular interest to our study. For each query token at index $\mathbf x_i \in \{1,\ldots,hw\}$ in $\mathbf Q^l_i$, $\mathbf A_{i,j}^l(\mathbf x_i) \in \mathbb R^{hw}$ represents a probability distribution of attention weights over all key tokens in $\mathbf{K}_j^l$.
This mechanism is particularly important as it enables each spatial location in one view to aggregate information from any location in other views. For our analysis and our method, we compute the normalized cross-view attention map \(\mathbf{A}^{l}_{i,j} = \texttt{softmax} (\mathbf{Q}^{l}_i(\mathbf{K}^{l}_j)^\top / \sqrt d)\) for each view pair \((i,j)\) with \(i \ne j\).

\begin{figure}[ht]
  \centering
  \includegraphics[width=1\linewidth]{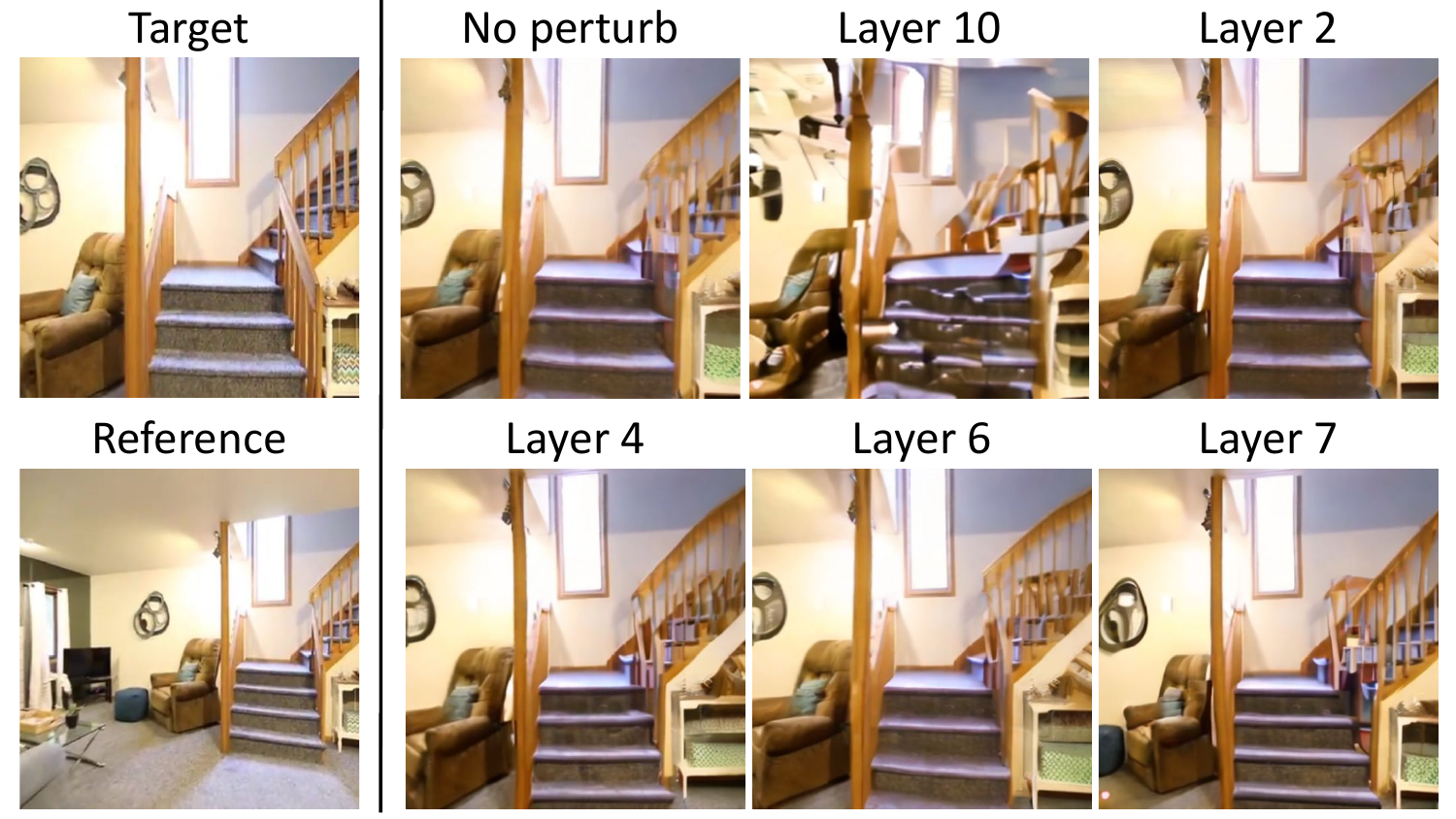}

  \caption{\textbf{Effect of layer-wise attention perturbation.} Following the perturbation procedure of PAG~\cite{pag}, perturbing earlier layers barely changes generation quality, while perturbing layer 10 collapses geometric consistency and severely degrades quality.}
  \label{fig:perturb}
  \vspace{-0.8em}
\end{figure}
\subsection{Motivation and analysis}
\label{sec:analysis}

\begin{figure*}[ht]
  \centering
  \vspace{-10px}
  \begin{subfigure}[b]{0.33\linewidth}
    \includegraphics[width=\linewidth,trim=12 8 12 6,clip]{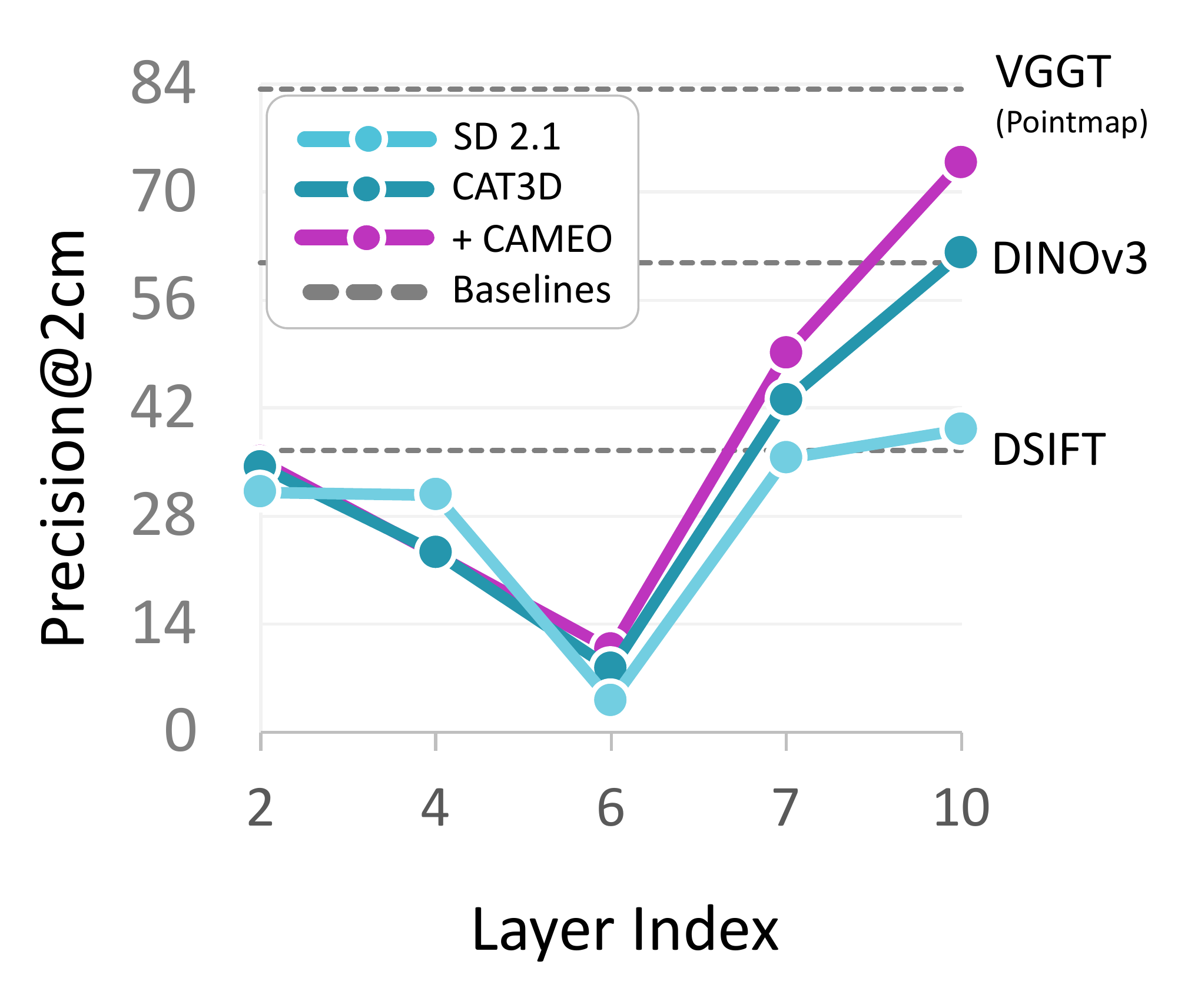}
    \caption{Precision across different layers}
    \label{fig:main_analysis_a}
  \end{subfigure}\hfill
  \hspace{-5px}
    \begin{subfigure}[b]{0.33\linewidth}
    \includegraphics[width=\linewidth,trim=12 8 12 6,clip]{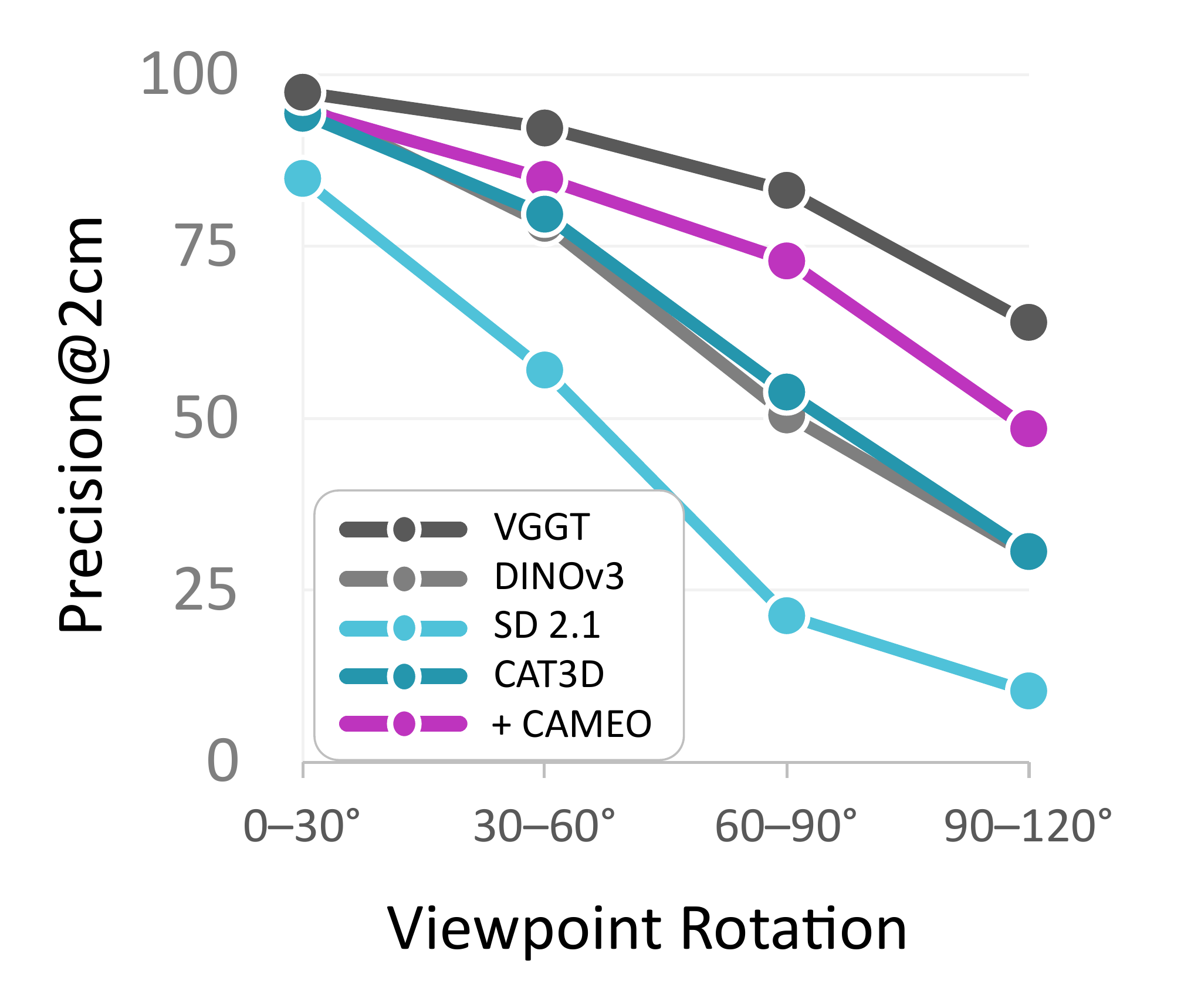}
    \caption{Precision across relative viewpoint bins} % (b)
    \label{fig:main_analysis_b}
  \end{subfigure}
  \hspace{-5px}
  \begin{subfigure}[b]{0.33\linewidth}
    \includegraphics[width=\linewidth,trim=12 8 12 6,clip]{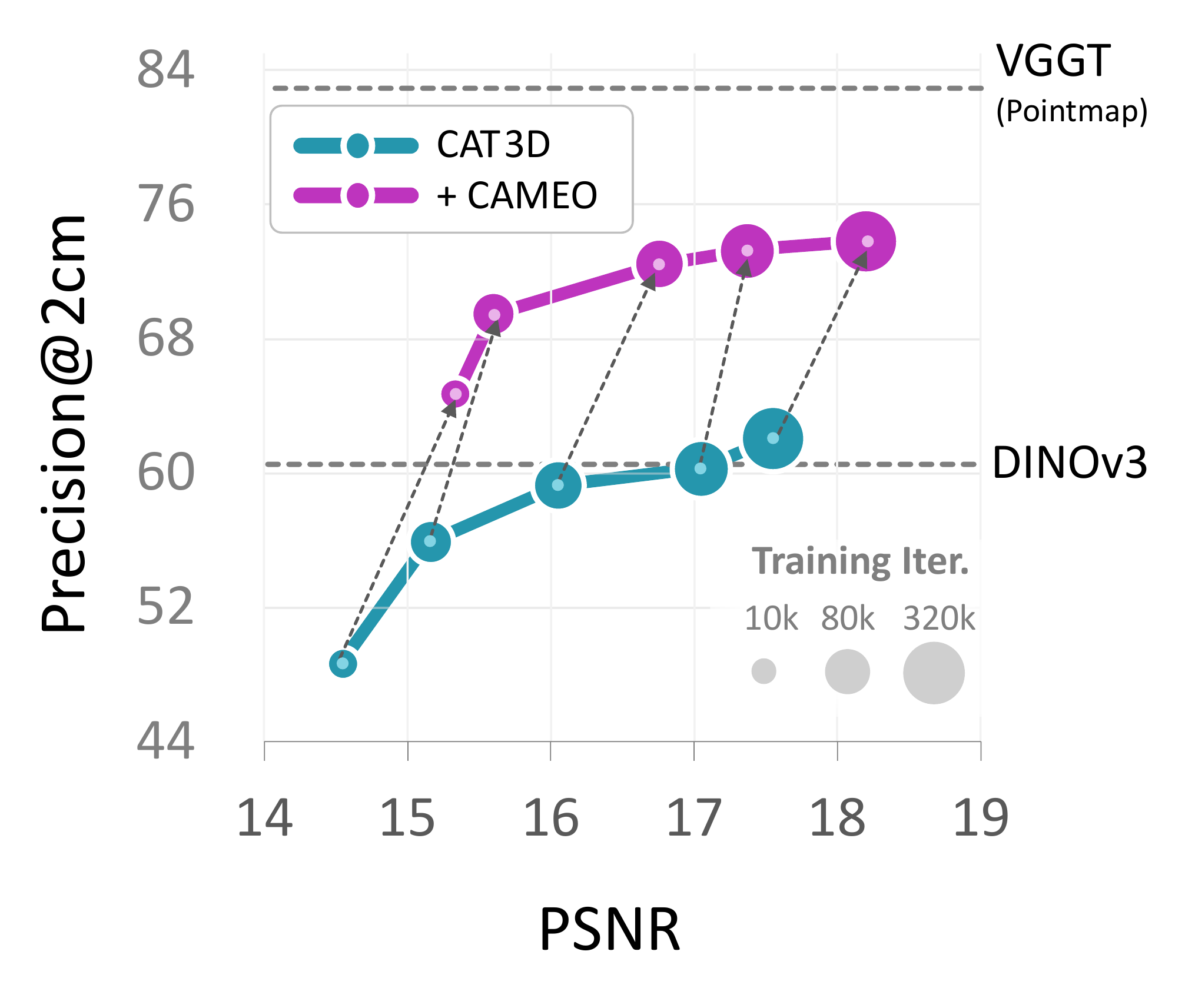}
    \caption{Precision and PSNR across training iterations} % (c)
    \label{fig:main_analysis_c}
  \end{subfigure}
  \vspace{-5pt}
  \caption{\textbf{Analysis of geometric correspondence in attention maps of the multi-view diffusion model~\cite{cat3d}.} (a) Correspondence precision across attention layers (\(l = 2, 4, 6, 7, 10\)), with other baselines~\cite{ldm,5551153,wang2025vggt,simeoni2025dinov3}. (b) The correspondence precision of layer $l=10$ with baselines, across viewpoint rotation. (c) The correspondence precision of layer $l=10$ improves during training. }
  \label{fig:main_analysis}
  \vspace{-15px}
\end{figure*}

We begin by qualitatively analyzing the layer-wise behavior of the cross-view attention maps. Specifically, given two images $\mathbf{I}_1$ (target) and $\mathbf{I}_2$ (reference) of the same scene from different viewpoints, we first obtain query $\mathbf{Q}^l_1$ and key $\mathbf{K}^l_2$ during the denoising process to compute the cross-view attention map $\mathbf{A}_{1,2}^{l}$. For each query token index $\mathbf{x_1}$ in $\mathbf{Q}_1$, we identify the key token index with the highest attention weight. As shown in~\cref{fig:layerwise}, layer $l=10$ exhibits a consistent pattern: a query token consistently attends to its geometrically corresponding point in another view. This observation leads us to hypothesize that the cross-view attention maps in the model capture geometric correspondence. We provide the visualization results for all layers in~\cref{sec:appendix-analysis_layerwise}.

To rigorously investigate and understand this emergent behavior, we quantitatively measure the geometric correspondence encoded in the attention maps~\cite{an2025cross}. Following Probe3D~\cite{el2024probing}, which proposed a framework to evaluate semantic or geometric correspondence, we evaluate the geometric correspondence within the cross-view attention map of the multi-view diffusion model~\cite{cat3d}. Specifically, given an image pair $\mathbf{I}_1$ and $\mathbf{I}_2$ with a viewpoint rotation angle $\theta$ from the NAVI dataset~\cite{navi}, we estimate the correspondence for each query point of $\mathbf{I}_1$ with another image $\mathbf{I}_2$ by identifying the location with the highest attention weight in $\mathbf{A}_{1,2}^{l}$. We also evaluate the baselines, including (i) the attention map of CAT3D before finetuning, (SD2.1~\cite{ldm}-initialized, denoted SD2.1), (ii)
the dense SIFT descriptors (DSIFT) from SIFT Flow~\cite{5551153}, (iii) the intermediate features of DINOv3-L~\cite{simeoni2025dinov3}, and (iv) the point maps from VGGT~\cite{wang2025vggt}. We summarize the evaluation results in~\cref{fig:main_analysis} and provide details in~\cref{sec:appendix-analysis_detail}.

\paragrapht{Emergence of geometric correspondence in attention layers.}
As shown in~\cref{fig:main_analysis_a}, we report the geometric correspondence precision averaged over all viewpoint rotations~($\theta$) for layers $l = 2,4,6,7,10$, which are the first layers of each U-Net block. Compared to SD2.1~\cite{ldm}, CAT3D~\cite{cat3d} develops significantly stronger correspondence in its cross-view attention maps in layers $l=7,10$ that outperform DSIFT~\cite{5551153}. In particular, layer $l=10$ achieves precision comparable to DINOv3-L~\cite{simeoni2025dinov3}. For $l=10$, the precision for each viewpoint rotation angle is given in~\cref{fig:main_analysis_b}. For small viewpoint rotation, $\theta = 0^{\circ} - 30^{\circ}$, CAT3D achieves as high a precision as VGGT~\cite{wang2025vggt} pointmaps, demonstrating its ability to capture geometric correspondence.

\paragrapht{Geometric correspondence improves throughout training.}
In~\cref{fig:main_analysis_c}, we plot the geometric correspondence precision in $l=10$ across training iterations in CAT3D~\cite{cat3d}. We observe that both the correspondence precision and PSNR increase monotonically during training, demonstrating that the model progressively learns to encode more accurate geometric correspondence in the attention map. This positive correlation between correspondence precision and generation quality suggests that this correspondence underpins the synthesis quality. We further verify this in~\cref{fig:perturb} by perturbing its attention map. Following PAG~\cite{pag}, we force the 3D self-attention at a given layer to an identity mapping so that each query only attends to its corresponding identity location. Perturbing earlier layers (e.g., $l=2,4,6,7$) leaves the outputs nearly unchanged, whereas perturbing $l=10$ collapses the scene into a heavily distorted, geometrically implausible image. This demonstrates that the ability of $l=10$  to capture geometric correspondences across views is crucial for view-consistent generation.

\paragrapht{Denoising objectives provide limited correspondence supervision.}
Although the attention layers capture geometric correspondence and their precision correlates with the generation quality, a substantial performance gap remains compared to a strong baseline, VGGT~\cite{wang2025vggt}. As shown in~\cref{fig:main_analysis_a,fig:main_analysis_c}, CAT3D's attention maps exhibit significantly lower correspondence precision than VGGT, even after extensive training iterations. More critically,~\cref{fig:main_analysis_b} reveals that CAT3D's attention maps struggle to capture accurate geometric correspondence under large viewpoint rotations, while VGGT maintains relatively robust performance across varying viewpoints. This suggests that the standard denoising objective alone is insufficient for the model to learn accurate geometric correspondence, motivating our exploration of explicit correspondence supervision.

\subsection{CAMEO: Correspondence-attention alignment for multi-view diffusion models}
\label{sec:method}
We propose CAMEO, correspondence-attention alignment for multi-view diffusion models, which explicitly supervises the attention maps with the geometric correspondence. Our analysis (\cref{sec:analysis}) shows that layer $l=10$ captures the strongest geometric correspondence and that its precision correlates with generation quality, yet it degrades significantly under large viewpoint rotation, indicating room for improvement. By directly supervising $l=10$ with geometric correspondence, we enable faster learning of geometric relationships and improved novel view synthesis.

\paragrapht{Geometric correspondence map.}
Given a set of images \(\{\mathbf{I}_i\}_{i=1}^F\), dense geometric correspondence identifies matching pixels between any image pair $(\mathbf I_i,\mathbf I_j)_{i\ne j}$ such that the matched pixels correspond to the same point in the 3D space. These correspondences can be aligned to the token-level resolution $h\times w$ through spatial downsampling or interpolation. Specifically, the geometric correspondences for all $\mathbf x_i$ can be defined as $\{(\mathbf x_i, \mathbf x_j), \mathbf M_{i,j}(\mathbf x_i)\}^{hw}$, where $\mathbf x_i, \mathbf x_j \in \{1,\ldots,hw\}$ are the query token indices and $\mathbf M_{i,j}(\mathbf{x}_i) \in[0,1]$ is a visibility mask, indicating whether the corresponding point of $\mathbf x_i$ is visible in $\mathbf I_j$.

Then we build a one-hot correspondence vector $\mathbf P_{i,j}(\mathbf x_i)\in \mathbb R^{hw}$ for each $\mathbf x_i$, where its $\mathbf x_j$-th element is $1$ while others are all zero. We then stack these vectors over all query token indices $\mathbf x_i$ to get a geometric correspondence map $\mathbf P_{i,j}\in \mathbb R^{hw \times hw}$. We construct such maps for all pairs among the $N$ reference and $M$ target images. 

We follow DUSt3R~\cite{wang2024dust3r} to obtain the correspondences from pointmaps by finding the nearest neighbor in 3D space. To compute $\mathbf{M}_{i,j}$, we verify cycle correspondence consistency. For a query token index \(\mathbf{x}_i\), we obtain the cycle correspondence index $\hat {\mathbf x}_i$ using correspondences $(\mathbf x_i, \mathbf x_j)$ and $(\mathbf x_j, \hat {\mathbf x}_i)$. We convert the flattened token indices to their corresponding 2D spatial coordinates and set \(\mathbf{M}_{i,j}(\mathbf{x}_i) = 1\) only when $\| \mathbf{p}(\mathbf{x}_i) - \mathbf{p}(\hat{\mathbf{x}}_i) \|_2 \le \tau$, where $\mathbf{p}(\cdot)$ denotes the mapping from token index to 2D spatial coordinate and $\tau$ is a cycle consistency threshold.

\paragrapht{Correspondence-attention alignment.}
Our method simply aligns the cross-view attention map $\mathbf{A}_{i,j}^{l} \in \mathbb{R}^{hw\times{hw}}$ with the geometric correspondence map $\mathbf{P}_{i,j}\in \mathbb{R}^{{hw}\times{hw}}$ for all view pairs and query tokens. In~\cref{fig:attention_and_corr_map}, we visualize the cross-view attention map $\mathbf{A}^{\,l}_{i,j}(\mathbf{x}_i)$ and geometric correspondence map $\mathbf P_{i,j}(\mathbf x_i)$ for a query token index $\mathbf x_i$ of each view $i$. In practice, multi-view diffusion models typically employ multi-head attention~\cite{3dim, zero123, zero123++, cat3d, mvgenmaster, bolt3d, stablevirtualcamera, matrix3d}, allowing different heads to capture diverse patterns and representations. Applying uniform alignment across all heads would restrict the architectural flexibility and limit the model's expressive capacity. To mitigate this, we employ a projection head on the attention logits before $\texttt{softmax}(\cdot)$, using a simple multilayer perceptron (MLP).

We define the CAMEO loss on a layer \(l\):
\begin{equation}
\scalebox{0.9}{$
\mathcal{L}_{\text{CAMEO}}
= \mathbb{E}_{(i,j),\,\mathbf{x}_i}\!\left[
   \mathbf{M}_{i,j}(\mathbf x_i) \odot
   \mathrm{CE}\!\left(
      \mathbf{A}^{\,l}_{i,j}(\mathbf{x}_i),\,
      \mathbf{P}_{i,j}(\mathbf x_i)
   \right)
\right]
$}
\end{equation} where $\mathrm{CE}(\cdot)$ is the cross-entropy loss, and $\odot$ denotes element-wise multiplication.

\paragrapht{Training objective.} 
Our final training objective combines the standard denoising score matching loss $\mathcal{L}_{\text{denoise}}$ used in diffusion models~\cite{ho2020denoising} with the proposed correspondence-attention alignment loss as $\mathcal{L}_{\text{total}} = \mathcal{L}_{\text{denoise}} + \lambda \mathcal{L}_{\text{CAMEO}}$, where $\lambda$ is a hyperparameter.

\begin{figure*}[t]
  \centering
  \vspace{-1em}

  \begin{subfigure}[t]{0.33\linewidth}
      \centering
      \includegraphics[width=\linewidth]{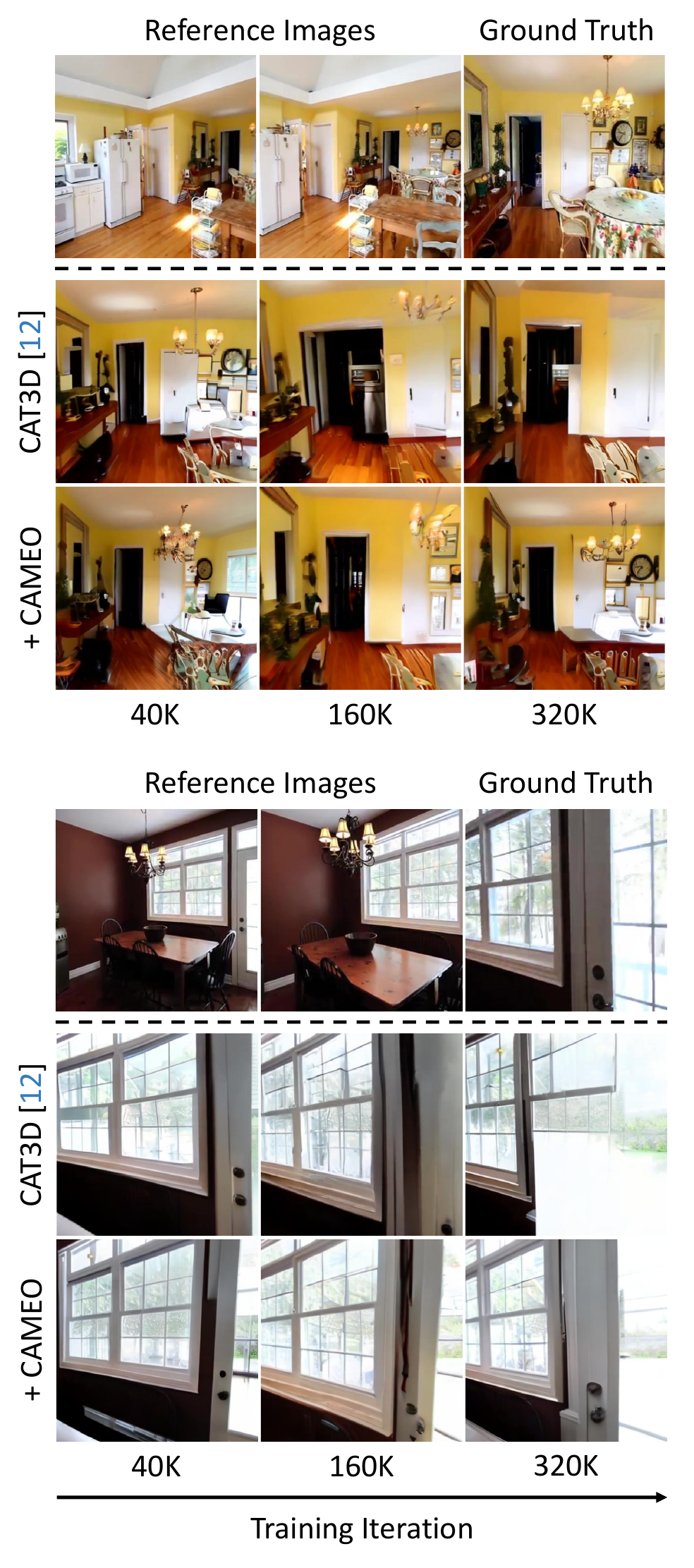}
      \caption{RealEstate10K~\cite{re10k}}
      \label{fig:main_qual_a}
  \end{subfigure}
  \hfill
  \begin{subfigure}[t]{0.33\linewidth}
      \centering
      \includegraphics[width=\linewidth]{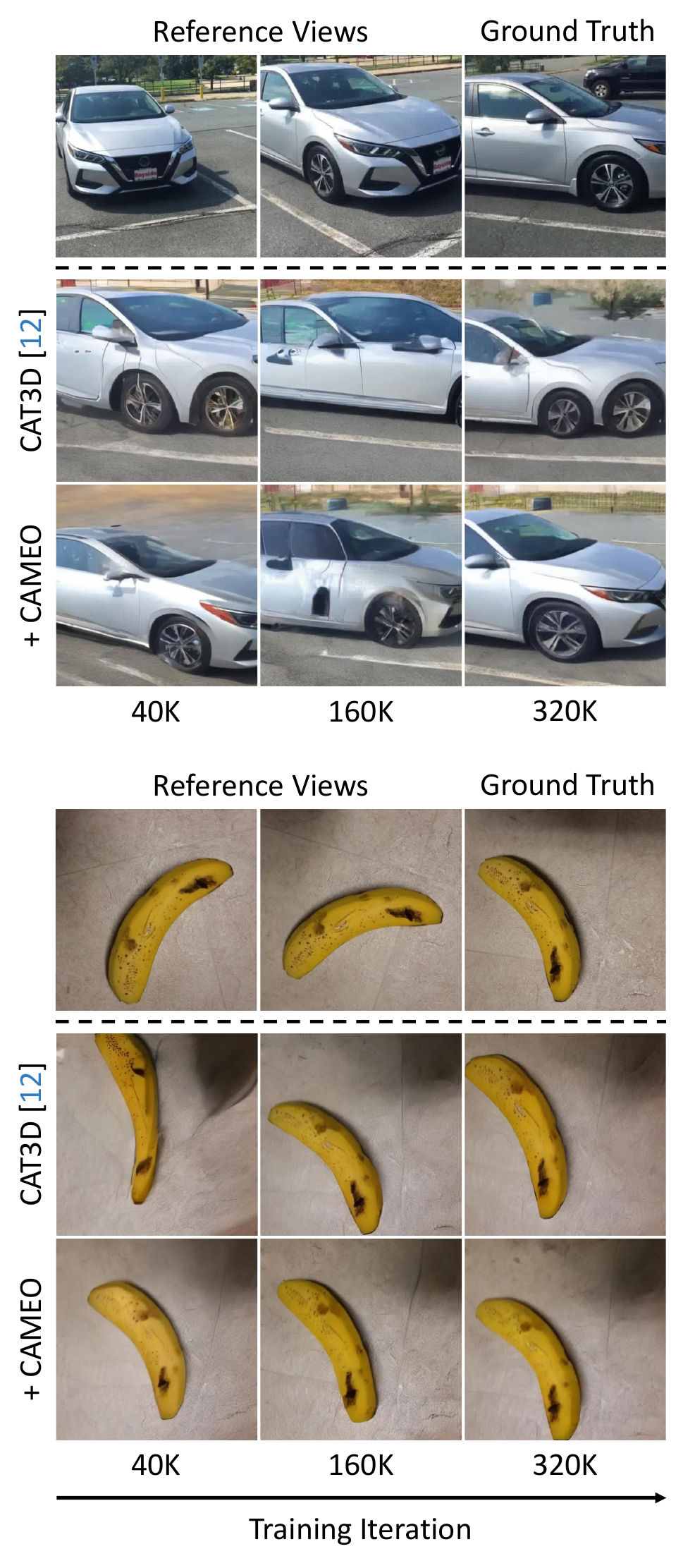}
      \caption{CO3D~\cite{reizenstein21co3d}}
      \label{fig:main_qual_b}
  \end{subfigure}
  \hfill
  \begin{subfigure}[t]{0.33\linewidth}
      \centering
      \includegraphics[width=\linewidth]{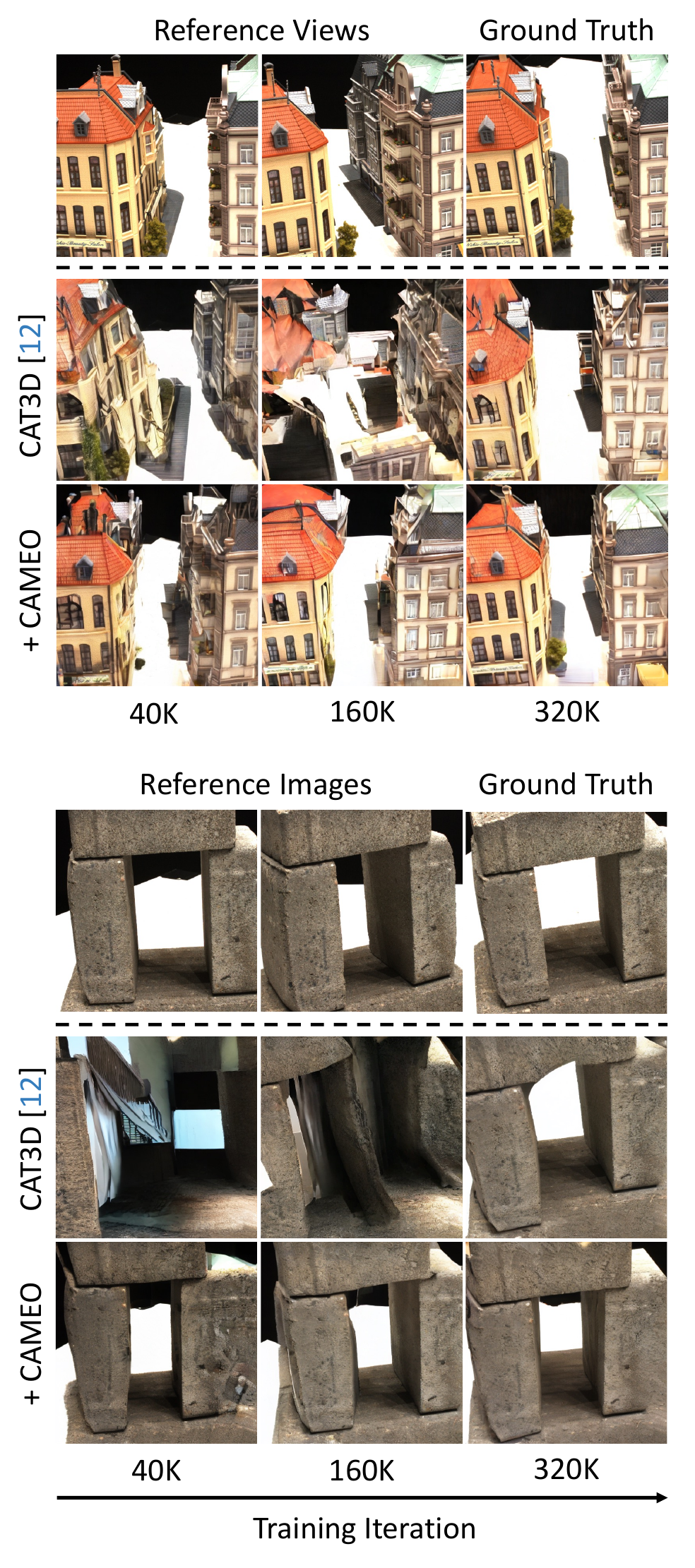}
      \caption{DTU~\cite{jensen2014largedtu} (Out-of-domain)}
      \label{fig:main_qual_c}
  \end{subfigure}

  \vspace{-0.75em}
  \caption{\textbf{Qualitative results} on (a) RealEstate10K~\cite{re10k}. (b) CO3D~\cite{reizenstein21co3d}, (c) DTU~\cite{jensen2014largedtu} (Out-of-domain).
    CAMEO accelerates learning of pose and geometric relationships compared to the baseline, as explicit correspondence supervision encourages geometric consistency and faster convergence in novel view synthesis. Additional results are provided in~\cref{sec:appendix-qualitative}.}
  \label{fig:main_qual}
  \vspace{-10pt}
\end{figure*}
\section{Experiments}

\label{sec:experiments}
We evaluate the effectiveness of \textbf{CAMEO} by addressing the following key questions:
\begin{itemize}[itemsep=0.25ex, topsep=1pt, leftmargin=*]
    \item Does CAMEO improve the quality of novel view synthesis under large viewpoint changes and complex scene?
    \item Can CAMEO accelerate the training of multi-view diffusion models?
    \item Does CAMEO remain effective under out-of-domain (OOD) settings, demonstrating generalization beyond the training distribution?
    \item Is CAMEO a general framework, which can be applied to any multi-view diffusion models?
\end{itemize}

\subsection{Setup}
\label{sec:implementation_details}
\textbf{Model.} We adopt CAT3D~\cite{cat3d} as the baseline multi-view diffusion model for our main experiments. Since the official implementation of CAT3D is not publicly available, we employ the re-implementation provided by MVGenMaster~\cite{mvgenmaster}. Following prior works~\cite{zero123,cat3d}, we initialize the model from pretrained Stable Diffusion 2.1 weights~\cite{ldm}. Additional architectural details are provided in~\cref{sec:appendix-cat3d_detail}. For comparison, we also train and evaluate two baseline methods: (1) REPA~\cite{repa}, a feature alignment method using self-supervised model~\cite{oquab2023dinov2} features, and (2) Geometry Forcing~\cite{wu2025geometry}, a feature alignment method that aligns diffusion features with geometric features extracted from a geometric prediction model~\cite{wang2025vggt}.

To verify that CAMEO generalizes beyond our baseline architecture~\cite{cat3d}, we further implement CAMEO in two additional architectures: MVGenMaster~\cite{mvgenmaster}, a state-of-the-art multi-view diffusion model that adopts geometric conditioning, and a DiT-based~\cite{li2024hunyuandit} multi-view diffusion model. Implementation details for these models are provided in~\cref{sec:appendix-implement_othermodel}.

\begin{table*}[t]
  \centering
  \scriptsize
  \setlength{\tabcolsep}{6pt}
  \renewcommand{\arraystretch}{1.05}
  \caption{\textbf{Novel view synthesis evaluation} on 
  RealEstate10K~\citep{re10k}, CO3D~\citep{reizenstein21co3d}, and DTU~\citep{jensen2014largedtu}.
  \textbf{Bold} numbers indicate the best within each iteration group.}
    \begin{tabular}{@{}cl|c|ccc|ccc|ccc@{}}
    \toprule
    & & & \multicolumn{3}{c}{\textbf{RealEstate10K}~\cite{re10k}} 
      & \multicolumn{3}{c}{\textbf{CO3D}~\cite{reizenstein21co3d}}
      & \multicolumn{3}{c}{\textbf{DTU~\cite{jensen2014largedtu} (Out-of-domain)}} \\
    \cmidrule(lr){4-6} \cmidrule(lr){7-9} \cmidrule(lr){10-12}
    {} & \textbf{Model} & Iter.
      & PSNR $\uparrow$ & SSIM $\uparrow$ & LPIPS $\downarrow$
      & PSNR $\uparrow$ & SSIM $\uparrow$ & LPIPS $\downarrow$
      & PSNR $\uparrow$ & SSIM $\uparrow$ & LPIPS $\downarrow$ \\
    \midrule

    % ----------------------- 10k -----------------------
    & CAT3D~\cite{cat3d} & \multirow{4}{*}{10k}
      & 16.68 & 0.617 & 0.377
      & 14.55 & \textbf{0.555} & 0.560
      & 9.35  & 0.285 & 0.603 \\
      
    & w/ REPA~\cite{repa} &
      & 16.82
      & 0.613
      & 0.379
      & 14.24
      & 0.542
      & 0.558
      & 9.81
      & 0.296
      & 0.607 \\
      
    & w/ Geometry Forcing~\cite{wu2025geometry} &
     & 17.71
     & 0.648
     & 0.362
      & 14.11 & 0.521 & 0.549 
      & 9.47
      & 0.303
      & 0.604 \\
      
    & \textbf{w/ CAMEO (Ours)} &
      & \textbf{18.00} & \textbf{0.650} & \textbf{0.346}
      & \textbf{15.33} & 0.548 & \textbf{0.521}
      & \textbf{9.99}  & \textbf{0.311} & \textbf{0.589} \\
    \midrule

    % ----------------------- 40k -----------------------
    & CAT3D~\cite{cat3d} & \multirow{4}{*}{40k}
      & 18.13 & 0.650 & 0.330
      & 15.16 & 0.557 & 0.531
      & 9.90  & 0.308 & 0.603 \\
      
    & w/ REPA~\cite{repa} &
      & 18.27
      & 0.654
      & 0.325
      & \textbf{15.69}
      & 0.560
      & \textbf{0.507}
      & 10.45
      & 0.342
      & 0.580 \\
      
    & w/ Geometry Forcing~\cite{wu2025geometry} &
      & 18.49
      & \textbf{0.670}
      & 0.328
      & 15.10 & 0.529 & 0.512
      & 9.89
      & \textbf{0.345}
      & 0.598 \\
      
    & \textbf{w/ CAMEO (Ours)} &
      & \textbf{18.56}
      & 0.667
      & \textbf{0.315}
      & 15.60
      & \textbf{0.566}
      & 0.513
      & \textbf{10.76}
      & 0.322
      & \textbf{0.552} \\
    \midrule

    % ----------------------- 80k -----------------------
    & CAT3D~\cite{cat3d} & \multirow{4}{*}{80k}
      & 18.99 & 0.682 & 0.317
      & 16.05 & 0.570 & 0.495
      & 10.29 & 0.321 & 0.574 \\
      
    & w/ REPA~\cite{repa} &
      & 18.70
      & 0.672
      & 0.312
      & 16.10
      & 0.573
      & 0.495
      & 10.31
      & 0.353
      & 0.585 \\
      
    & w/ Geometry Forcing~\cite{wu2025geometry} &
      & 18.92
      & 0.676
      & 0.316
      & 15.94 & 0.563 & 0.484
      & 10.43
      & 0.344
      & 0.587 \\
      
    & \textbf{w/ CAMEO (Ours)} &
      & \textbf{19.40}
      & \textbf{0.690}
      & \textbf{0.301}
      & \textbf{16.76}
      & \textbf{0.589}
      & \textbf{0.478}
      & \textbf{11.45}
      & \textbf{0.387}
      & \textbf{0.531} \\
    \midrule

    % ----------------------- 160k -----------------------
    & CAT3D~\cite{cat3d} & \multirow{4}{*}{160k}
      & 19.28 & 0.686 & 0.300
      & 17.05 & 0.579 & \textbf{0.456}
      & 10.49 & 0.321 & 0.588 \\
    & w/ REPA~\cite{repa} &
      & 19.42    & 0.698  & 0.292
      & 16.78 & 0.584 & 0.468
      & 11.72 & \textbf{0.385} & \textbf{0.522} \\
    & w/ Geometry Forcing~\cite{wu2025geometry} &
      & 19.24 & 0.685 & 0.305
      & 16.91 & 0.574 & 0.458
      & 11.69 & 0.373 & 0.554 \\
    & \textbf{w/ CAMEO (Ours)} &
      & \textbf{19.51} & \textbf{0.699} & \textbf{0.288}
      & \textbf{17.37} & \textbf{0.591} & \textbf{0.456}
      & \textbf{12.06} & 0.353 & 0.526 \\
    \midrule

    % ----------------------- 320k -----------------------
    & CAT3D~\cite{cat3d} & \multirow{3}{*}{320k}
      & 19.88 & 0.702 & 0.287
      & 17.55 & 0.601 & 0.448
      & \textbf{12.23}& \textbf{0.401} & 0.524 \\
    & w/ REPA~\cite{repa} &
      & 19.76 & 0.702 & 0.286
      & 17.61 & 0.603 & 0.453
      & 11.71 & 0.365 & \textbf{0.521} \\
    & \textbf{w/ CAMEO (Ours)} &
      & \textbf{20.16} & \textbf{0.716} & \textbf{0.279}
      & \textbf{18.20} & \textbf{0.608} & \textbf{0.425}
      & 12.16 & 0.380 & 0.526 \\
    \midrule

    & CAT3D~\cite{cat3d} & \multirow{3}{*}{400k}
      & 19.64	& 0.696 & 0.293 
      & 17.06 & 0.580 & 0.446
      & 11.16	& 0.337	& 0.588 \\ 
    & w/ REPA~\cite{repa} &
      & 19.96 & 0.709 & 0.283
      & 17.09 & 0.580 & 0.445
      & 10.92 & 0.365 & 0.561 \\
    & \textbf{w/ CAMEO (Ours)} &
      & \textbf{20.07} & \textbf{0.710} & \textbf{0.277} 
      & \textbf{17.74} & \textbf{0.603} & \textbf{0.426}
      & \textbf{12.58} & \textbf{0.402} & \textbf{0.493} \\
    \bottomrule
\end{tabular}
  \label{tab:main}
  
\end{table*}

\begin{table}[b]
\centering
\scriptsize
\setlength{\tabcolsep}{3pt}
\renewcommand{\arraystretch}{1.05}
\caption{\textbf{Ablation studies of CAMEO} on RealEstate10K~\citep{re10k}. 
All at 40k iterations.}
\label{tab:ablation}

\begin{tabular}{@{}c c c c c c@{}}
  \toprule
  \textbf{Part} & \textbf{Factors} & \textbf{Variants} & PSNR$\uparrow$ & SSIM$\uparrow$ & LPIPS$\downarrow$ \\
  \midrule
\multirow{2}{*}{(a)} & \multirow{2}{*}{MLP head} & $\times$   
& 18.08 & 0.653 & 0.343 \\ & 
& \checkmark & \textbf{18.31} & \textbf{0.658} & \textbf{0.337} \\
\midrule
\multirow{3}{*}{(b)} & \multirow{3}{*}{Loss weight ($\lambda$)} 
& 0.01          & 18.22 & 0.657 & 0.351 \\&
& \textbf{0.02} & 18.31 & \textbf{0.658} & \textbf{0.337} \\&
& 0.03          & \textbf{18.37} & 0.656 & 0.377 \\
\midrule
\multirow{2}{*}{(c)} & \multirow{2}{*}{Loss type}
& L1 & 17.84 & 0.641 & 0.342 \\ &
& \textbf{$\mathbf{CE}(\cdot)$} & \textbf{18.31} & \textbf{0.658} & \textbf{0.337} \\
\midrule
\multirow{3}{*}{(d)} & \multirow{3}{*}{Consistency threshold ($\tau$)}
& $\infty$ & 18.18 & 0.656 & 0.341 \\ &
& 3 & 17.49 & 0.648 & 0.338 \\ &
& \textbf{1.5} & \textbf{18.31} & \textbf{0.658} & \textbf{0.337} \\
\bottomrule
\end{tabular}
\end{table}

\paragrapht{Dataset.} 
For the main experiments, we train the models on the scene-level (RealEstate10K~\cite{re10k}) and object-centric (CO3D~\cite{reizenstein21co3d}) datasets separately to demonstrate that our method is applicable to both scene-level and object-level synthesis. Each training sample consists of $F=4$ views, where 1 to 3 views are randomly masked as target views while the rest as references.
For the main evaluation, we measure the performance under two settings: one reference view with three target views (1-to-3) and two reference views with two target views (2-to-2), covering a diverse range of camera poses. We randomly sample 280 scenes from the RealEstate10K test set~\cite{re10k} and 240 scenes from the CO3D test set~\cite{reizenstein21co3d}, evaluating models trained on their respective datasets. For an out-of-domain (OOD) evaluation, we evaluate the scene-level model on the validation split of the DTU dataset (object-centric)~\cite{jensen2014largedtu}, processed by MVSplat~\cite{chen2024mvsplat}, and we conduct evaluation under 2-to-2 view setting. All models are trained and evaluated at 512$\times$512 resolution.

\paragrapht{Geometric correspondence map.}
We use off-the-shelf geometry model~\cite{wang2025vggt} to get pointmaps and bilinearly interpolate them to compute token-level geometric correspondence. We provide the details, including computation cost in~\cref{sec:appendix-implement_corr_map}.

\paragrapht{Implementation details.} 
In the main experiments, we apply CAMEO to layer $l = 10$, which shows the strongest emergent correspondence among all layers, yet the correspondence degrades significantly under large viewpoint rotations (\cref{sec:analysis}). We set a loss weight $\lambda = 0.02$, and the cycle consistency threshold $\tau = 1.5$. For REPA~\cite{repa}, we apply the REPA loss $\mathcal{L}_{\text{REPA}}$ to $l=3$ with a loss weight of 0.5, following the observation in the original paper. For Geometry Forcing~\cite{wu2025geometry}, we apply the angular alignment loss $\mathcal{L}_{\text{Angular}}$ and the scale alignment loss $\mathcal{L}_{\text{Scale}}$ at layer $l=3$ using VGGT~\cite{wang2025vggt} features, with a loss weight of 0.5.

For MVGenMaster~\cite{mvgenmaster}, we apply CAMEO at layer $l=10$ since it has the same UNet architecture as CAT3D~\cite{cat3d}. For the DiT-based model~\cite{li2024hunyuandit}, we apply CAMEO at $l=32$, identified by our analysis as the layer capturing the strongest geometric correspondence. We provide the detailed analysis of the DiT-based model in~\cref{sec:appendix-analysis-generalizability}.

\paragrapht{Training details.} We keep the batch size to 6 and train models with AdamW optimizer~\cite{LoshchilovH19}, adopting a fixed learning rate of 2.5e-5 and a weight decay of 0.01. Following ~\cite{cat3d,mvgenmaster}, we apply classifier-free guidance (CFG)~\cite{dhariwal2021diffusion} training by randomly dropping camera condition with a probability of 0.1. At inference, we use the DDIM sampler~\cite{song2020denoising} with 50 sampling steps and CFG with a weight of 2.0. All experiments are conducted on 2 NVIDIA A100 (40GB) GPUs.

\subsection{Main results}
\label{sec: exp detail}
\paragrapht{Training efficiency.} 
To investigate how CAMEO influences the training dynamics of multi-view diffusion models, we compare CAMEO with the baseline at intermediate training steps. As shown in~\cref{tab:main}, our method achieves faster convergence than the baseline~\cite{cat3d} and REPA~\cite{repa}. Specifically, CAMEO reaches a PSNR above 19.4 at 80k iterations, whereas the baseline requires 160k or more iterations to achieve the same performance — corresponding to a 2$\times$ acceleration. These results demonstrate that CAMEO enables more efficient learning of geometric structure in multi-view diffusion models.

\paragrapht{Novel view synthesis quality.}
The benefits of CAMEO extend beyond training efficiency to improvements in the final quality of novel view synthesis. As shown in~\cref{tab:main}, CAMEO surpasses both the baseline~\cite{cat3d}, REPA~\cite{repa}, and Geometry Forcing~\cite{wu2025geometry} at nearly all intermediate training iterations. These results indicate that aligning attention to a specific layer yields larger gains than feature-level alignment. Importantly, CAMEO consistently outperforms the baseline after convergence (beyond 320k iterations), demonstrating that it not only accelerates training but also improves final performance. 
Furthermore,~\cref{fig:main_qual} demonstrates that CAMEO produces novel views that are more aligned with the ground-truth images and better preserve object structure compared to the baseline. Specifically, in~\cref{fig:main_qual_c}, CAMEO captures the overall geometric structure of the building and brick as early as 40k iterations, whereas the baseline fails to recover these appearances. Even after 320k iterations, the baseline~\cite{cat3d} still struggles to accurately generate the building's roof and brick details, while CAMEO generates them with high geometric consistency. These results confirm that explicit correspondence supervision significantly improves overall NVS quality. Additional qualitative examples are provided in~\cref{sec:appendix-qualitative}.

\paragrapht{Generalization to OOD setting.} The advantages of CAMEO are not limited to in-domain settings. As shown in~\cref{tab:main}, even when evaluated on the object-centric DTU dataset~\cite{jensen2014largedtu}, our method consistently outperforms the baseline. This suggests that CAMEO enables the model to learn a general geometric understanding that extends beyond the training distribution.
\begin{table*}[t]
  \centering
  \begin{minipage}[t]{0.25\textwidth}
    \centering
    \scriptsize
    \setlength{\tabcolsep}{3pt}
    \renewcommand{\arraystretch}{1.0}
    \caption{\textbf{Layer Ablation.} \textbf{Bold} numbers indicate the best within each iteration group.}
    \label{tab:layer_ablation}

\begin{tabular}{@{}c | c | c c c@{}}
\toprule
Layer & Iter. &
PSNR$\uparrow$ & SSIM$\uparrow$ & LPIPS$\downarrow$ \\
\midrule

% 40k block
2            & \multirow{5}{*}{40k} & 18.26 & \textbf{0.664} & \textbf{0.336} \\
4            &                      & 17.98 & 0.653          & 0.343          \\
6            &                      & 17.84 & 0.645          & 0.342          \\
7            &                      & 18.07 & 0.648          & 0.339          \\
\textbf{10}  &                      & \textbf{18.31} & 0.658 & 0.337 \\
\midrule

% 80k block
2            & \multirow{5}{*}{80k} & 18.80& 0.676& \textbf{0.315}\\
4            &                      & 18.78& 0.673& 0.320\\
6            &                      & 18.19& 0.663& 0.326\\
7            &                      & 18.60& 0.666& 0.323\\
\textbf{10}  &                      & \textbf{19.08}&\textbf{ 0.681}& 0.316\\
\bottomrule
\end{tabular}
\end{minipage}
\hfill
  % ------------ Left table (MVGenMaster) ------------
  \begin{minipage}[t]{0.35\textwidth}
    \centering
  % ------------ Right table (Hunyuan-DiT) ------------
    \scriptsize
    \setlength{\tabcolsep}{3pt}
    \renewcommand{\arraystretch}{1.05}

    \caption{\textbf{CAMEO on MVGenMaster~\cite{mvgenmaster}.} \textbf{Bold} numbers indicate the best within each iteration group.}
    \label{tab:sota_mvgenmaster}

    \begin{tabular}{@{}cl|c|ccc@{}}
      \toprule
      {} & \textbf{Model} & Iter. & PSNR$\uparrow$ & SSIM$\uparrow$ & LPIPS$\downarrow$ \\
      \midrule
      & MVGenMaster~\cite{mvgenmaster} & \multirow{2}{*}{ 20k }
        & 17.35 & 0.649 & 0.327 \\
      & \textbf{w/ CAMEO (Ours)} &
        & \textbf{18.31} & \textbf{0.671} & \textbf{0.315} \\
      \midrule

      & MVGenMaster~\cite{mvgenmaster} & \multirow{2}{*}{ 40k }
        & 18.64 & 0.682 & 0.306 \\
      & \textbf{w/ CAMEO (Ours)} &
        & \textbf{18.70} & \textbf{0.668} & \textbf{0.305} \\
      \midrule

      & MVGenMaster~\cite{mvgenmaster} & \multirow{2}{*}{ 60k }
        & 18.84 & 0.678 & 0.305 \\
      & \textbf{w/ CAMEO (Ours)} &
        & \textbf{19.21} & \textbf{0.695} & \textbf{0.304} \\
        \midrule
      & MVGenMaster~\cite{mvgenmaster} & \multirow{2}{*}{ 120k }
        & 19.45 & 0.699 & 0.295 \\
      & \textbf{w/ CAMEO (Ours)} &
        & \textbf{19.56} & \textbf{0.700} & \textbf{0.292} \\

      \bottomrule
    \end{tabular}
  \end{minipage}
  \hfill
  % ------------ Right table (Hunyuan-DiT) ------------
  \begin{minipage}[t]{0.35\textwidth}
    \centering
    \scriptsize
    \setlength{\tabcolsep}{3pt}
    \renewcommand{\arraystretch}{1.05}

    \caption{\textbf{CAMEO on Hunyuan-DiT~\cite{li2024hunyuandit}} \textbf{Bold} numbers indicate the best within each iteration group.}
    \label{tab:sota_hunyuan}

    \begin{tabular}{@{}cl|c|ccc@{}}
      \toprule
      {} & \textbf{Model} & Iter. & PSNR$\uparrow$ & SSIM$\uparrow$ & LPIPS$\downarrow$ \\
      \midrule
      & Hunyuan-DiT~\cite{li2024hunyuandit} & \multirow{2}{*}{ 20k }
        & 14.40 & 0.533 & 0.459 \\
      & \textbf{w/ CAMEO (Ours)} &
        & \textbf{16.17} & \textbf{0.575} & \textbf{0.373} \\
      \midrule

      & Hunyuan-DiT~\cite{li2024hunyuandit} & \multirow{2}{*}{ 40k }
        & 17.00 & 0.594 & 0.321 \\
      & \textbf{w/ CAMEO (Ours)} &
        & \textbf{17.57} & \textbf{0.612} & \textbf{0.294} \\
      \midrule

      & Hunyuan-DiT~\cite{li2024hunyuandit} & \multirow{2}{*}{ 60k }
        & 17.55 & 0.610 & 0.289 \\
      & \textbf{w/ CAMEO (Ours)} &
        & \textbf{18.52} & \textbf{0.639} & \textbf{0.260} \\
        \midrule
      & Hunyuan-DiT~\cite{li2024hunyuandit} & \multirow{2}{*}{ 120k }
        & 19.30 & 0.661 & 0.218 \\
      & \textbf{w/ CAMEO (Ours)} &
        & \textbf{19.75} & \textbf{0.677} & \textbf{0.211} \\
    
      \bottomrule
    \end{tabular}
  \end{minipage}

\end{table*}

\paragrapht{Generalization to other architectures.}
We also evaluate CAMEO on the other baseline architectures, a state-of-the-art~\cite{mvgenmaster} and a DiT-based model~\cite{li2024hunyuandit}. As shown in~\cref{tab:sota_hunyuan,tab:sota_mvgenmaster}, CAMEO yields improvements in PSNR, SSIM, and LPIPS across both models, particularly in early training iterations. These results demonstrate that our framework improves multi-view diffusion models with varying architectures, confirming its model-agnostic nature. We provide qualitative results of both models in~\cref{sec:appendix-qualitative}.

\subsection{Ablation studies}
To analyze the contribution of each component in CAMEO, we conduct comprehensive ablation studies by systematically varying its core components: the alignment layer, the presence of an MLP head, the weighting parameter $\lambda$, the loss function, and the cycle consistency threshold $\tau$. All models are trained on the RealEstate10K dataset~\cite{re10k} with a batch size of 3 and evaluated in a 2-to-2 view setting.

\cref{tab:layer_ablation} presents our alignment layer analysis. While supervising intermediate layers ($l=4, 6, 7$) yields poor performance, layer $l=10$ demonstrates competitive performance with $l=2$ at 40k iterations and clearly outperforms it by 80k iterations, validating our analysis in~\cref{sec:analysis}.

As shown in~\cref{tab:ablation}, incorporating an MLP head proves superior to direct alignment, as it preserves the representational diversity of multi-head attention. We find that $\lambda=0.02$ (balancing loss magnitudes) and $\tau=1.5$ (ensuring reliable matches) yield optimal results. Furthermore, cross-entropy loss achieves superior performance compared to L1 loss, as it directly aligns attention probability distributions.

\subsection{Analysis}
We analyze the effect of CAMEO by measuring correspondence precision and visualizing the attention maps of CAT3D~\cite{cat3d} trained with and without CAMEO.

\paragrapht{Correspondence accuracy.} ~\cref{fig:main_analysis} presents two observations. In~\cref{fig:main_analysis_a,fig:main_analysis_b}, CAMEO increases attention correspondence precision across all viewpoint rotations, even surpassing feature matching in DINOv3~\cite{simeoni2025dinov3}. As shown in~\cref{fig:main_analysis_c}, CAMEO can push the baseline~\cite{cat3d} to achieve both higher correspondence precision and PSNR at the same iterations, indicating faster learning of geometric correspondence and earlier gains in generation quality.

\begin{figure}[t]
  \centering
  \includegraphics[width=\linewidth]{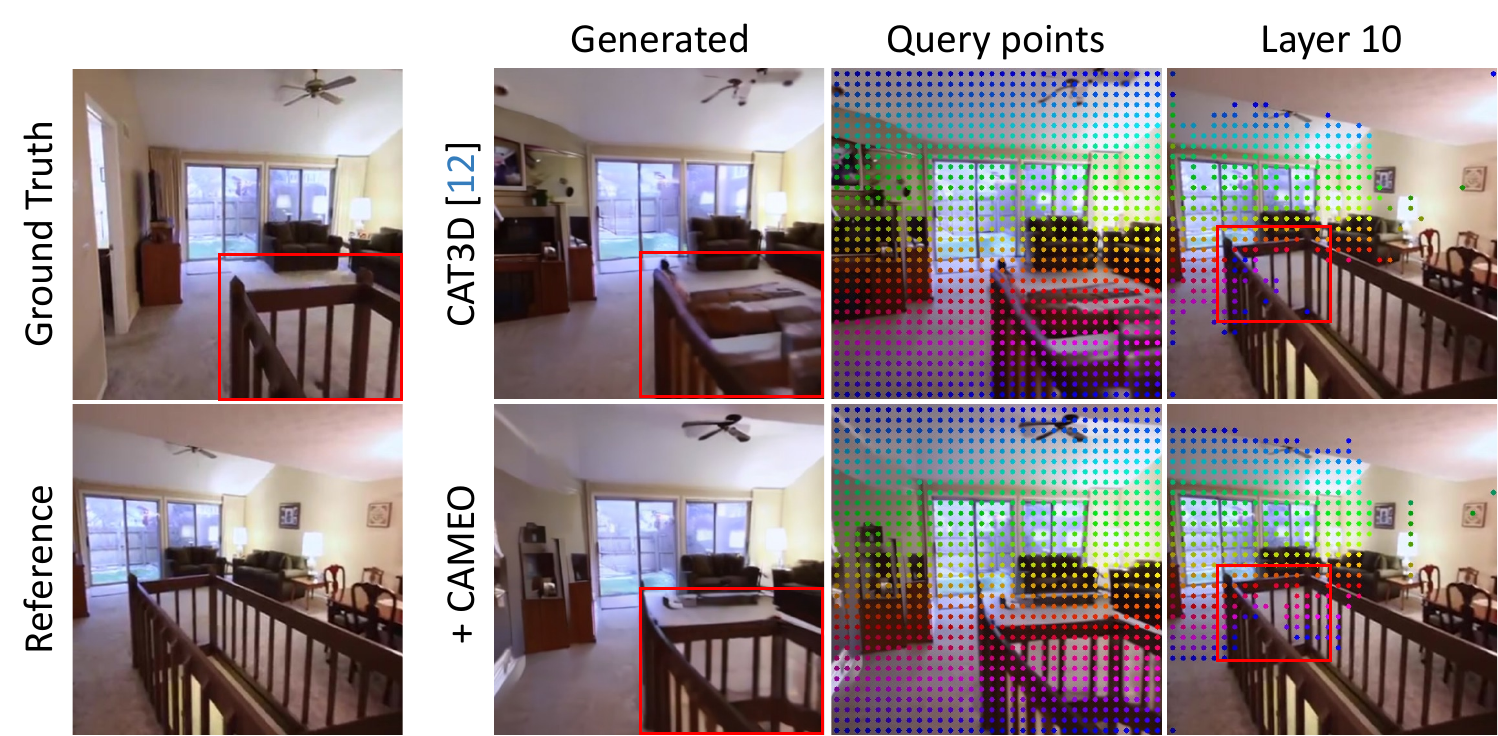}
  \caption{\textbf{Correspondence analysis in $l=10$.} In CAT3D~\cite{cat3d}, pink query points on the handrail fail to attend to their geometric counterparts in the reference, whereas CAMEO succeeds. As a result, the handrail is accurately generated only in CAMEO.
}
  \vspace{-10pt}
  \label{fig:attn_after}
  % \vspace{-10pt}
\end{figure}

\paragrapht{Qualitative analysis.} In~\cref{fig:attn_after}, CAT3D~\cite{cat3d} produces a distorted handrail, while CAMEO preserves the correct shape with fine detail and accuracy. The attention maps reveal the mechanism behind this performance gap. For CAT3D~\cite{cat3d}, query points on the handrail fail to attend to the handrail region in the reference image. In contrast, CAMEO correctly attends to the corresponding handrail region, resulting in precise geometric reconstruction with structural details. This demonstrates that CAMEO successfully guides the model to learn more accurate geometric correspondences, which directly improves novel view synthesis performance.
\section{Conclusion}
In this work, we presented an analysis of multi-view diffusion models, revealing that their 3D self-attention maps learn emergent geometric correspondence that is critical for generation quality. We also identified the limitations of this implicit signal, which proved fragile under large viewpoint changes. Building on these findings, we introduced CAMEO, a simple yet effective technique that injects explicit geometric supervision into the attention layers. We demonstrated that our approach significantly improves novel view synthesis quality, enhances geometric fidelity under challenging viewpoints, and accelerates the training. CAMEO is model-agnostic and can be readily integrated into existing and future architectures that employ cross-view attention. We hope our findings on the link between the correspondence and the attention inspire further research in geometry-aware generative modeling.

\clearpage

\clearpage
\setcounter{section}{0}
\renewcommand{\thesection}{\Alph{section}}

\section*{\Large Appendix}
This appendix presents additional experimental results and further details of our proposed method, CAMEO.
\begin{itemize}
    \item \cref{sec:appendix-diffusion} reviews the fundamentals of diffusion models. 
    \item \cref{sec:appendix-cat3d_detail} describes the architecture of multi-view diffusion models in detail. 
    \item \cref{sec:appendix-analysis} provides a detailed analysis of correspondence in multi-view diffusion models, including the analysis setup. 
    \item \cref{sec:appendix-implement_detail} covers implementation details, including correspondence map derivation, and implementation of baseline models.
    \item \cref{sec:appendix:abl_featcost} shows additional ablation studies.
    \item \cref{sec:appendix-qualitative} presents additional qualitative results. 
    \item \cref{sec:appendix-3drecon} provides the results and implementation details of 3D reconstruction.
    \item \cref{sec:appendix-limitation} describes the limitations of CAMEO.
    \item \cref{sec:appendix-futurework} discusses future directions.
\end{itemize}

\section{Preliminaries for diffusion models}
\label{sec:appendix-diffusion}
\label{secA}

Diffusion models~\cite{ho2020denoising, song2020denoising} are a class of generative models that learn data distributions by reversing a gradual noising process. Starting from clean data samples $x_0 \sim p_{\text{data}}(x)$, a forward process incrementally corrupts them with Gaussian noise to produce a sequence of latent variables $\{x_t\}_{t=1}^T$. A neural network is then trained to approximate the reverse process, progressively denoising a sample from pure Gaussian noise back into a realistic data point.

\paragrapht{Denoising diffusion probabilistic models.} Denoising Diffusion Probabilistic Models (DDPM)~\cite{ho2020denoising} define a forward noising process $q(x_t|x_{t-1})$ with a variance schedule $\{\beta_t\}_{t=1}^T$, where $\alpha_t = 1 - \beta_t$ and $\bar{\alpha}_t = \prod_{s=1}^t \alpha_s$. At an arbitrary timestep $t$, the closed form of the noising process is
\begin{equation}
x_t = \sqrt{\bar{\alpha}_t}\, x_0 + \sqrt{1 - \bar{\alpha}_t}\,\epsilon, \quad \epsilon \sim \mathcal{N}(0, I).
\end{equation}
The generative task is to learn the reverse process $p_\theta(x_{t-1}|x_t)$ such that a sample from $x_T \sim \mathcal{N}(0, I)$ can be gradually denoised to yield $x_0 \sim p_{\text{data}}$. In practice, this reverse transition is parameterized by a neural network $\epsilon_\theta(x_t, t)$ that predicts the noise, leading to
\begin{equation}
\scalebox{0.9}{$
p_\theta(x_{t-1}|x_t) := \mathcal{N}\!\left(x_{t-1}; \frac{1}{\sqrt{\alpha_t}}\Big(x_t - \frac{\beta_t}{\sqrt{1 - \bar{\alpha}_t}}\,\epsilon_\theta(x_t, t)\Big),\, \sigma_t^2 I\right),
$}
\end{equation}
where $\sigma_t^2$ can be fixed or learned. Training is performed with the denoising objective
\begin{equation}
\mathcal{L}_{\text{denoise}}(\theta) = \mathbb{E}_{x_0,\epsilon,t}\big[\|\epsilon - \epsilon_\theta(x_t, t)\|_2^2\big],
\end{equation}
which corresponds to score matching~\cite{hyvarinen2005estimation}, since $\epsilon_\theta(x_t, t)$ approximates the score function $-\sigma_t \nabla_{x_t} \log p(x_t)$. Moreover, by reparameterization one can directly obtain an estimate of the clean sample $x_0$ at timestep $t$ as
\begin{equation}
\hat{x}_0(x_t) = \frac{1}{\sqrt{\bar{\alpha}_t}}\Big(x_t - \sqrt{1 - \bar{\alpha}_t}\,\epsilon_\theta(x_t, t)\Big),
\end{equation}
which provides an explicit reconstruction of the data from noisy inputs and plays a key role in both DDPM sampling and extensions such as DDIM.

\paragrapht{Denoising diffusion implicit models.} Denoising Diffusion Implicit Models (DDIM)~\cite{song2020denoising} build upon DDPM but modify the formulation to allow for a deterministic, non-Markovian sampling procedure that substantially accelerates generation. Instead of requiring $T$ iterative reverse steps, DDIM introduces a reparameterized reverse process where the current latent $x_t$ can be deterministically mapped to $x_{t-1}$ using both the predicted clean image $\hat{x}_0(x_t)$ and the predicted noise $\epsilon_\theta(x_t, t)$. Specifically, the reverse update is
\begin{equation}
x_{t-1} = \sqrt{\bar{\alpha}_{t-1}}\, \hat{x}_0(x_t) + \sqrt{1 - \bar{\alpha}_{t-1}}\, \epsilon_\theta(x_t, t).
\end{equation}

This deterministic formulation allows skipping intermediate steps in the reverse trajectory without retraining the model, leading to fast sampling while preserving high generative quality. DDIM thus serves as a practical alternative to DDPM and is widely adopted in applications where efficient and scalable generation is crucial.

\section{Details of the multi-view diffusion model}
\label{sec:appendix-cat3d_detail}
\label{secB}

\begin{figure*}[ht]
  \centering
  \vspace{-10px}
  \makebox[\linewidth][c]{
    \includegraphics[width=1.0\linewidth]{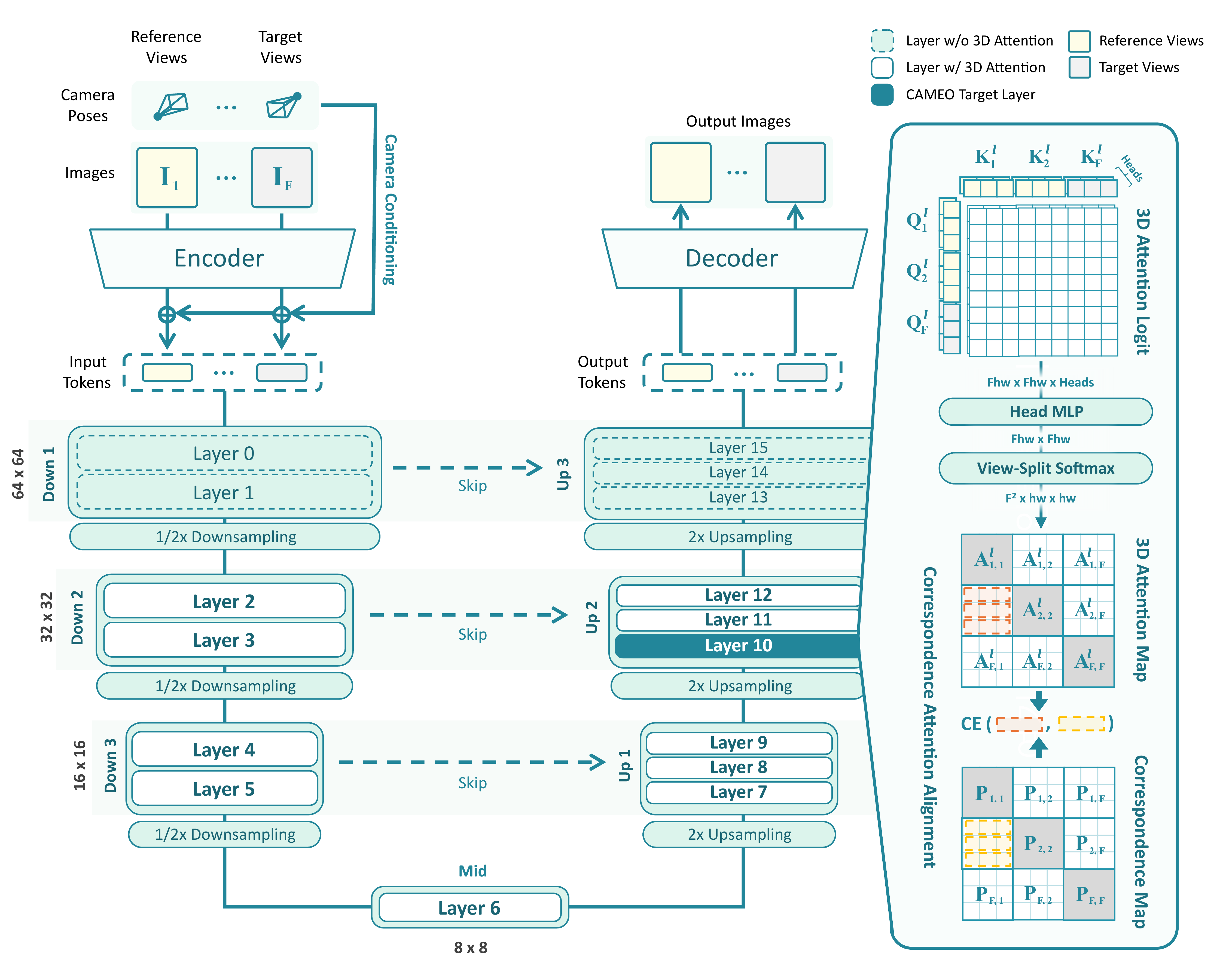}
  }
  \vspace{-15px}
\caption{\textbf{Model architecture of CAT3D~\cite{cat3d} with our proposed CAMEO framework.} While maintaining the original architecture, CAMEO introduces an additional correspondence-attention alignment loss, $L_{\text{CAMEO}}$, at the target layer (Layer 10) to supervise the attention map. Note that the visibility mask $\mathbf{M}_{i,j}$, which filters out occluded or unreliable correspondences via 3D cycle consistency, is omitted in this visualization for simplicity.}
  \label{fig:cat3d}
\end{figure*}

\paragraph{Implementation.} 
Our baseline model is CAT3D~\cite{cat3d}, a multi-view extension of Stable Diffusion 2.1~\cite{ldm}. CAT3D adapts the latent text-to-image diffusion framework by inflating the 2D self-attention layers into 3D self-attention, enabling interactions across different views. Although the official implementation and model weights of CAT3D are not publicly available, we adopt the reproduction provided by MVGenMaster~\cite{mvgenmaster}, which faithfully replicates CAT3D's training and evaluation pipeline. 

\paragrapht{Network architecture.} 
The underlying architecture consists of three downsampling blocks, one mid-block, and three upsampling blocks. Each downsampling block contains two layers, the mid-block contains one layer, and each upsampling block contains three layers. Each layer comprises a spatial convolution followed by a self-attention module.

In CAT3D, standard self-attention layers are replaced with inflated 3D self-attention layers to capture inter-view dependencies. This 3D attention is applied in all blocks except the first and last (\ie, it is implemented in downsampling blocks 2 \& 3, the mid-block, and upsampling blocks 1 \& 2). In total, there are 11 inflated 3D self-attention layers used in our analysis.

The input images of resolution \(512 \times 512\) are encoded by the VAE encoder into latent features of size \(64 \times 64\). Gaussian noise is added to the target latents for generation, while the reference latents remain unchanged. To form the conditioning latent, we first compute the Plücker ray embedding~\cite{xu2023dmv3d}, which encodes per-pixel camera rays, and concatenate it with a binary visibility mask indicating the reference images. This conditioning signal is then passed through a shallow convolutional network to match the dimensionality of the image latents. Finally, the conditioning latents are added to the image latents, producing the multi-view input representation for the diffusion U-Net.

Each downsampling block reduces the spatial resolution by a factor of 2, producing feature maps of size \(32 \times 32\), \(16 \times 16\), and \(8 \times 8\), respectively. The mid-block operates at the lowest resolution of \(8 \times 8\). The upsampling blocks then progressively restore the spatial resolution back to \(16 \times 16\), \(32 \times 32\), and \(64 \times 64\). Finally, the latent is passed through the VAE decoder to reconstruct the full-resolution image of size \(512 \times 512\).

\section{Detailed analysis}
\label{secC}
\label{sec:appendix-analysis}
In~\cref{sec:analysis}, we report the attention analysis for layer $l=2,4,6,7,10$, which are the first layers of each block. In~\cref{sec:appendix-analysis_layerwise}, we provide the qualitative analysis for all attention layers $l=2-12$. In~\cref{sec:appendix-analysis_detail}, we provide details of the correspondence precision measured in the NAVI dataset~\cite{navi} and report the precision for all attention layers. In~\cref{sec:appendix-analysis-generalizability}, we conduct a fine-grained analysis of layer-wise geometric correspondence in the diffusion transformer~\cite{li2024hunyuandit} based multi-view diffusion model to validate the generalizability of our systematic layer selection protocol.

\subsection{Qualitative analysis}
\label{sec:appendix-analysis_layerwise}
\begin{figure*}
    \centering
    \includegraphics[width=0.9\linewidth]{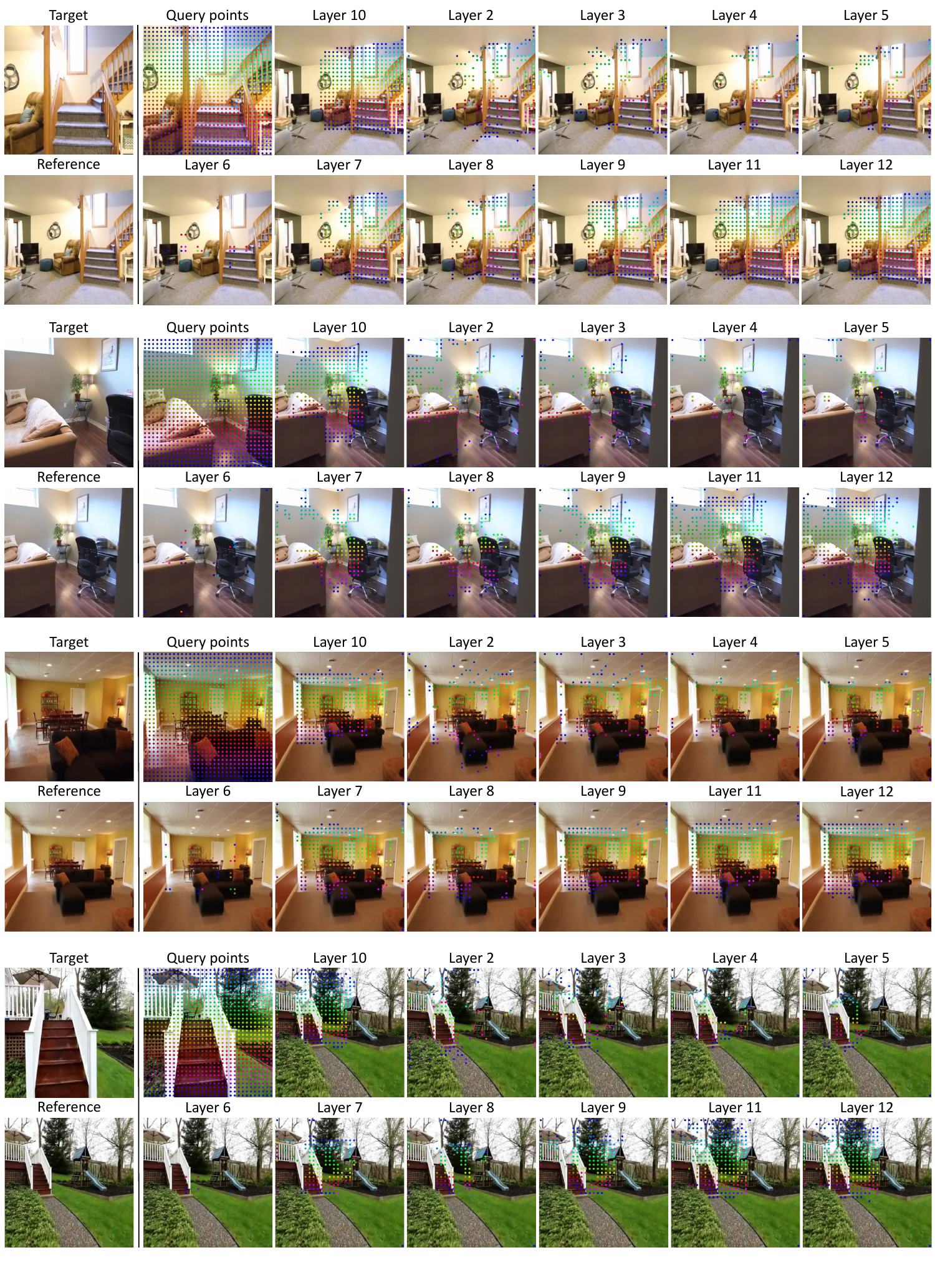}
    \caption{\textbf{Layer-wise behavior of the multi-view diffusion model (CAT3D~\cite{cat3d})'s attention map.} For each query point on the target image, model's maximum attending point in the reference image is marked with the same color as the query point.}
    \label{appendix:analysis-grid_query}
\end{figure*}
\paragrapht{Layer-wise behavior.} As shown in~\cref{appendix:analysis-grid_query}, the attention map in layer $l=10$ of CAT3D~\cite{cat3d} consistently attends to geometrically corresponding points relative to the query points. Similarly, layer $l=11,12$ also attend to geometrically corresponding points, while earlier layers ($l=2-6$) do not. Layers $l=7-9$ exhibit similar behavior, but their corresponding points are sparse and noisy compared to those in deeper layers ($l=10,12$). This layer-wise behavior demonstrates that attention maps in certain layers learn to capture geometric correspondence.

\begin{figure*}[t]
    \centering
    \includegraphics[width=0.88\linewidth]{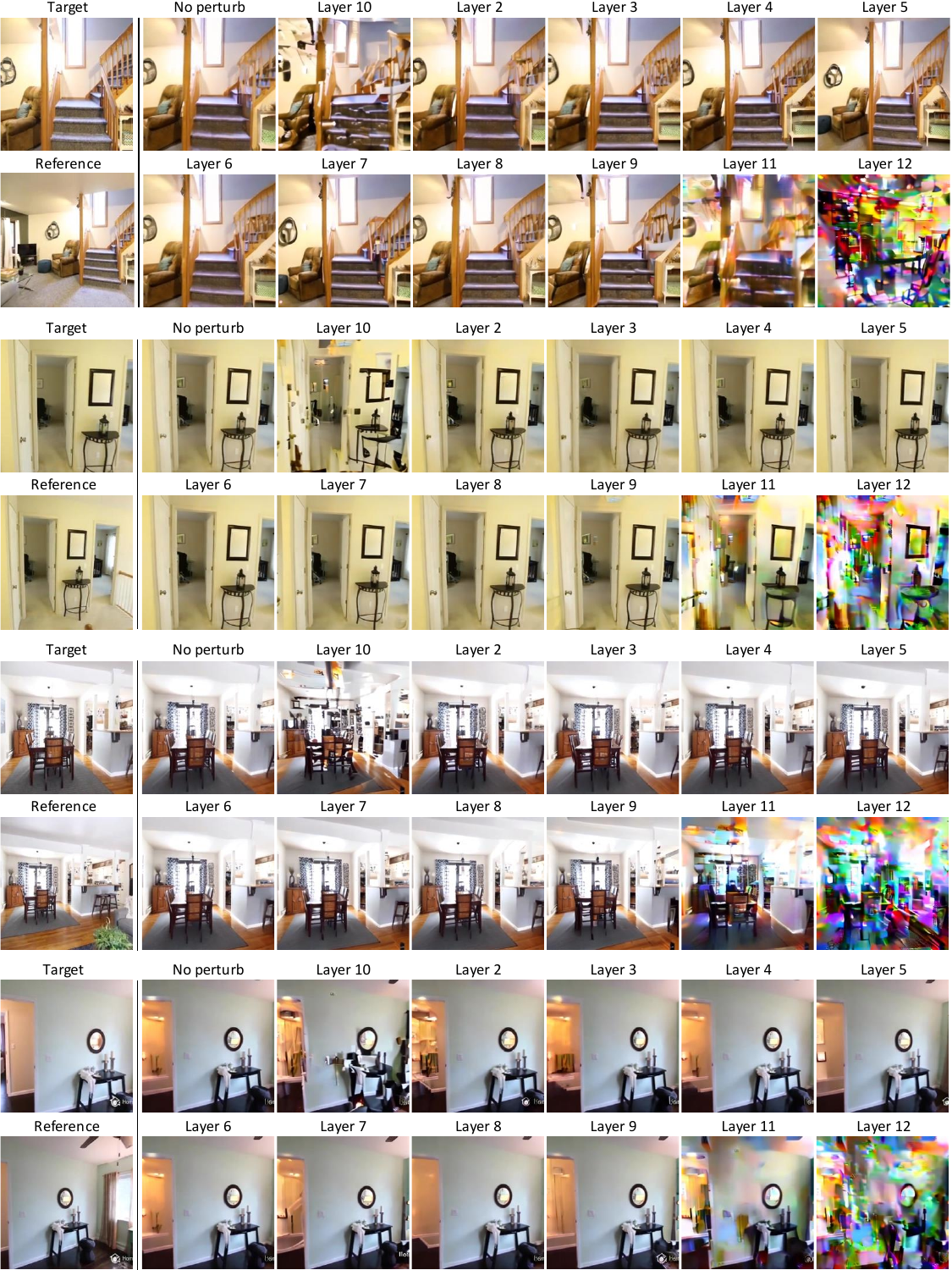}
    \caption{\textbf{Layer-wise perturbation results.} Perturbing earlier layers barely changes generation quality, while perturbing layer 10 collapses geometric consistency and severely degrades quality.}
    \label{appendix:analysis-perturb}
\end{figure*}
\paragrapht{Perturbation analysis.} Following~\cite{pag,kim2025seg4diff}, we perturb the 3D self-attention maps of CAT3D~\cite{cat3d} by enforcing the map into an identity mapping, where each query token attends exclusively to itself. The generation results in~\cref{appendix:analysis-perturb} demonstrate that perturbing layer $l=10$ causes severe structural fragmentation. In contrast, perturbing deeper layers ($l=11, 12$) results in color distortion or degradation but preserves the structural layout (\eg, the tables and plaque remain recognizable). Notably, perturbing other layers has a negligible impact on the generated images. This implies that the geometric correspondence captured specifically at layer $l=10$ is fundamental to maintaining the structural consistency of the generated view.

\subsection{Correspondence estimation}
\label{sec:appendix-analysis_detail}
\paragraph{Dataset.} We evaluate geometric correspondence on the NAVI Dataset~\cite{navi}, which consists of 36 object scans with the ground-truth 3D geometry. The dataset provides high-quality assets, including intrinsics, extrinsics, depth maps, and object masks, enabling the inference of dense pixel-level correspondences. While Probe3D~\cite{el2024probing} utilized the \textit{wild} split (same object, different backgrounds) of the dataset, we employ the \textit{multiview} split to align with the Novel View Synthesis (NVS) setting, where the background remains consistent across views. To construct the evaluation set, we subsample 25\% of the object views and select image pairs where the relative rotation $\theta$ of the destination view is within $120^\circ$. This results in a total of 245 source-destination pairs. 

\paragraph{Procedure.} 
Given two images $\mathbf{I}_1$ and $\mathbf{I}_2$ of the same scene, we aim to identify pixel pairs corresponding to the same 3D surface point.
We first extract a grid of feature descriptors from each image using the corresponding backbone, and resize this grid to a spatial resolution of 128 $\times$ 128.
Correspondences are then estimated by matching all feature descriptors from $\mathbf{I}_1$ to its nearest neighbor in $\mathbf{I}_2$ based on a pre-defined distance function.
To mitigate inaccurate matches, we apply Lowe's ratio test~\cite{lowe2004distinctive}, filtering for unique matches by comparing the distances to the first and second nearest neighbors.
For each feature token $p$, let $q_0$ and $q_1$ be the first and second nearest neighbors, respectively.
We compute the ratio $r$ as
\begin{equation}
r = 1 - \frac{D(p, q_0)}{D(p, q_1)},
\end{equation}
where        $D(x, y)$ denotes the distance between two features: cosine distance for feature-based descriptors, and $\ell_2$ distance in the 3D coordinate space for pointmap representations.
We retain the top $1000$ matches with the highest ratio $r$ for evaluation.

\paragraph{Evaluation.} 
We evaluate correspondence precision using the 3D Euclidean distance between the reprojected points. Given a correspondence pair, we back-project both image points into a common 3D space using ground-truth depth and camera parameters. A match is deemed correct if the 3D distance between the two points is below a threshold of $\rho = 2\mathrm{cm}$. Note that we denote this metric as \textit{Precision} instead of \textit{Recall} (as used in Probe3D~\cite{el2024probing}) to accurately reflect our protocol of evaluating a fixed number of top-1,000 candidates.

In \cref{fig:main_analysis_b}, we report the per-bin Precision@2cm averaged over samples in each angular bin. In the following section, we provide the detailed setup for correspondence extraction in CAT3D~\cite{cat3d}, SD2.1~\cite{ldm}, DINOv3~\cite{simeoni2025dinov3}, and VGGT pointmap~\cite{wang2025vggt}, Dense SIFT~\cite{5551153}.

\begin{itemize}
\item \textbf{CAT3D.} For the analysis, we evaluate CAT3D~\cite{cat3d} trained on the object-centric dataset CO3D~\cite{reizenstein21co3d}. We consider $\mathbf{I}_1$ and $\mathbf{I}_2$ as target and reference views. And we perform inference without camera conditioning and extract correspondences in a noise-free setting, \ie, without injecting additional diffusion noise. Query $\mathbf{Q}_1^l$ and key $\mathbf{K}_2^l$ descriptors are extracted at each layer $l$ (32 $\times$ 32 to 8 $\times$ 8 resolution).

\item \textbf{SD2.1.} We aim to analyze the correspondence in the attention map before finetuning. Therefore, using CAT3D~\cite{cat3d} architecture, we initialize the model weights with SD2.1 and measure the correspondence in the attention map. 

\item \textbf{DINOv3.} DINOv3~\cite{simeoni2025dinov3} is a state-of-the-art visual foundation model used across many downstream tasks, such as image retrieval, semantic segmentation, and dense matching. It is known for producing reliable patch-level matches.
We extract patch embeddings on a 32 $\times$ 32 grid and compute cosine similarity between patches across views to identify correspondences. We report the results for both ViT-B/16 and ViT-L/16 variants in~\cref{tab:suppl_navi}.

\item \textbf{VGGT pointmap.} We measure geometric correspondence based on $\ell_2$-norm nearest neighbor search. Since the pointmap explicitly encodes 3D coordinates at a resolution of 518 $\times$ 518, we use distance-based matching rather than cosine similarity. 

\item \textbf{Dense SIFT.}
For Dense SIFT, we follow the SIFT Flow~\cite{5551153} pipeline and compute dense SIFT descriptors on a multi-scale pyramid. We aggregate the descriptors onto a 128 $\times$ 128 grid and apply normalization following RootSIFT~\cite{arandjelovic2012three}. The resulting descriptors are used as feature tokens for cosine-distance matching, as in the other methods.

\end{itemize}
\begin{figure}[t!]
  \centering
    \includegraphics[width=\linewidth]{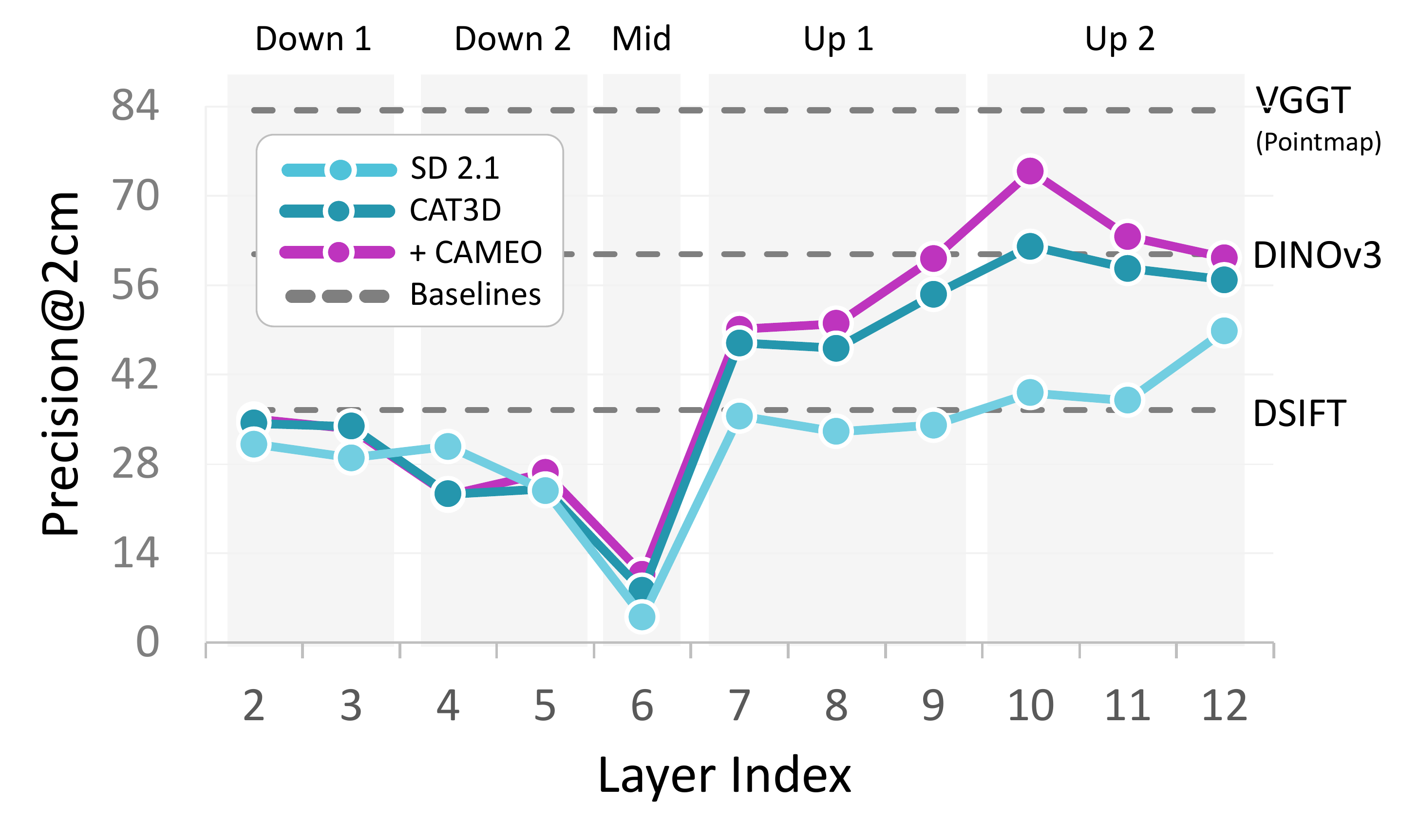}
    \vspace{-20px}
  \caption{\textbf{Analysis of geometric correspondence in all attention layers of the multi-view diffusion model~\cite{cat3d}.} Correspondence precision across all attention layers (\(l = 2-12\)), with other baselines~\cite{ldm,5551153,wang2025vggt,simeoni2025dinov3}. }
  \label{fig:appendix_layer_analysis}
  \vspace{-10px}
\end{figure}

\paragrapht{Correspondence precision.} In~\cref{fig:appendix_layer_analysis}, we present layer-wise correspondence precision across the U-Net architecture, with detailed quantitative results in~\cref{tab:suppl_navi}. In the downsampling blocks ($l=2-5$), SD2.1~\cite{ldm} and CAT3D~\cite{cat3d} exhibit comparable performance, both showing a sharp decline at the bottleneck ($l=6$) due to spatial compression. A clear divergence emerges in the upsampling blocks ($l=7-12$), where CAMEO demonstrates improved consistency over SD2.1~\cite{ldm}. Among all layers, $l=10$ shows the strongest correspondence, validating our analysis in~\cref{sec:analysis} that $l=10$ encodes the geometric correspondence.

With alignment, CAMEO consistently achieves higher correspondence precision than CAT3D~\cite{cat3d} across the upsampling blocks. Notably, even though our supervision is applied solely to the single target layer ($l=10$), performance gains are observed throughout the neighboring upsampling layers ($l=7-12$). This demonstrates that aligning one layer is sufficient to guide the model toward learning precise correspondences.

\subsection{Generalizability of layer selection}
\label{sec:appendix-analysis-generalizability}

In~\cref{sec:experiments}, we demonstrate the generalizability of CAMEO by applying it to the diffusion transformer (DiT) architecture~\cite{li2024hunyuandit}. In this section, we provide a fine-grained analysis of the layer-wise geometric correspondence in the DiT-based multi-view diffusion model to investigate the generalizability of systematic layer selection. Following the evaluation protocol in~\cref{sec:analysis}, we measure the geometric correspondence precision across all layers of the DiT model. As detailed in~\cref{tab:dit_navi}, we observe that layer $l=32$ exhibits the highest precision.

The layer-wise behavior of the DiT-based model differs from that of the UNet-based model~\cite{cat3d} due to their architectural differences. For instance, deeper layers ($l=35-39$) show low precision in geometric correspondence. However, despite these differences, the DiT-based model still captures geometric correspondence in specific layers, and CAMEO improves model performance by supervising it with our simple attention alignment method, as shown in~\cref{tab:sota_hunyuan}. These results demonstrate that CAMEO can be generalized to multi-view diffusion models with diverse architectures. Furthermore, our key finding—that multi-view diffusion models progressively learn to capture geometric correspondence in their attention maps during training—holds across different architectural designs.

\begin{figure*}[t]
  \centering
  \includegraphics[width=1\textwidth, height=0.32\textwidth]{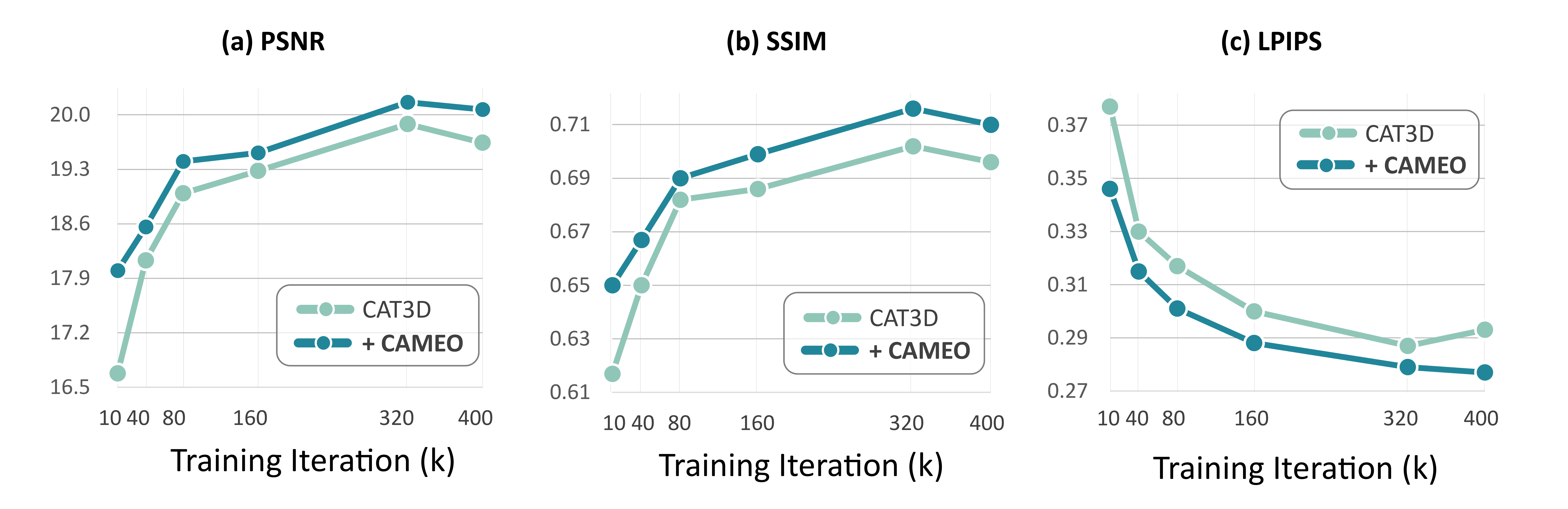}
  \vspace{-1.4em}
  \caption{\textbf{The relative improvements of CAMEO over CAT3D~\cite{cat3d} on RealEstate10K~\cite{re10k} dataset.} }
  % We report 2.0$\times$ times speed up in PSNR, LPIPS, and SSIM.}
  \label{fig:appendix_main_graph_re10k}
\end{figure*}
\begin{figure*}[t]
  \centering
  \includegraphics[width=1\textwidth, height=0.31\textwidth]{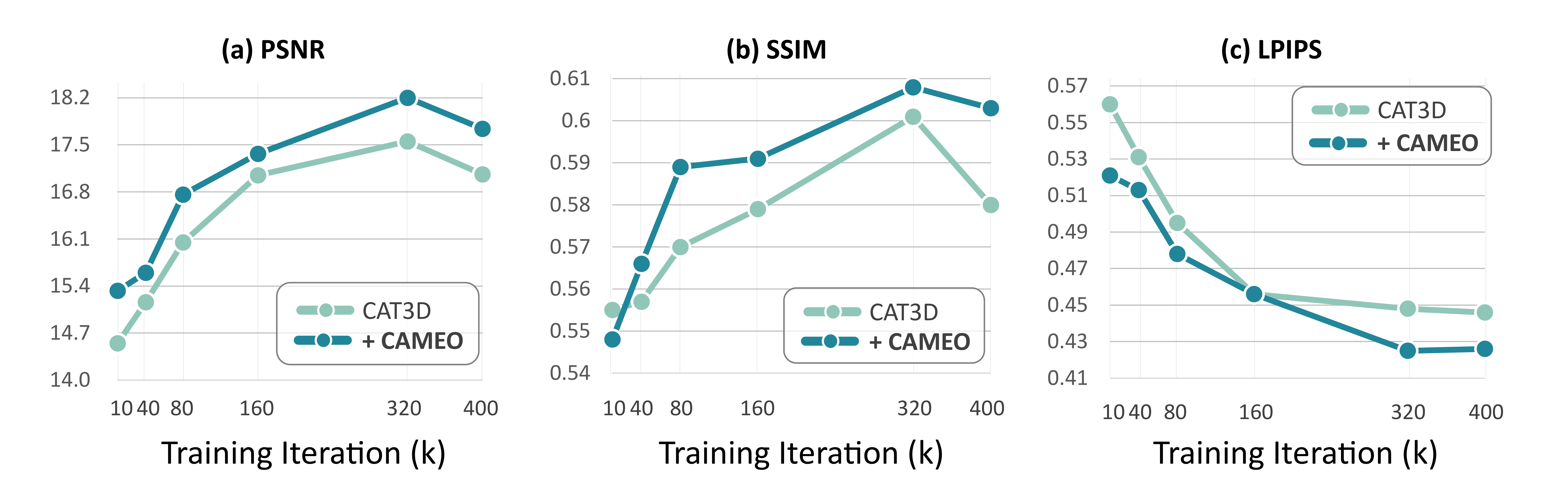}
  \vspace{-1.4em}
  \caption{\textbf{The relative improvements of CAMEO over CAT3D~\cite{cat3d} on CO3D~\cite{reizenstein21co3d} dataset.} }
  % We report 2.0$\times$ times speed up in PSNR, LPIPS, and SSIM.}
  \label{fig:appendix_main_graph_co3d}
  \vspace{-7px}

\end{figure*}

\section{Implementation details}
\label{secD}
\label{sec:appendix-implement_detail}
In~\cref{sec:appendix-eval_setting}, we provide the evaluation settings. In~\cref{sec:appendix-implement_corr_map}, we explain the detailed method to obtain correspondences from pointmaps, then evaluate the accuracy and efficiency. In \cref{sec:appendix-implement_othermodel}, we provide implementation details of a DiT-based~\cite{li2024hunyuandit} and a state-of-the-art multi-view diffusion model~\cite{mvgenmaster}.

\subsection{Evaluation settings} \label{sec:appendix-eval_setting} In~\cref{sec: exp detail}, we evaluate CAMEO against baselines~\cite{cat3d, repa} on the RealEstate10K~\cite{re10k}, CO3D~\cite{reizenstein21co3d}, and DTU~\cite{jensen2014largedtu} datasets. In the evaluation, to assess the model's robustness in maintaining view consistency under challenging scenarios, we curate a hard evaluation set. Specifically, for RealEstate10K~\cite{re10k} and CO3D~\cite{reizenstein21co3d}, we sample images that exhibit large viewpoint changes. While this aggressive sampling strategy naturally yields lower quantitative scores across all evaluated models (as seen in~\cref{tab:main}), it is essential for identifying failure cases in geometric consistency and rigorously assessing robustness under challenging scenarios.

\subsection{Correspondence from pointmap}
\label{sec:appendix-implement_corr_map} 
% In~\cref{sec:analysis,sec:method}, we use depth and camera parameters estimated by an off-the-shelf geometry model~\cite{wang2025vggt} to define pointmaps and correspondence. 

\paragrapht{Method.}
DUSt3R~\cite{wang2024dust3r} introduces an algorithm to establish pixel correspondences between two images through nearest neighbor search in 3D pointmap space. Specifically, for an image $\mathbf{I}_i \in \mathbb{R}^{H\times W \times 3}$, a pointmap $\mathbf{X}_i \in \mathbb{R}^{H\times W \times 3}$ represents the 3D coordinates of each pixel. Given an image pair $(\mathbf{I}_i,\mathbf{I}_j)$ where $i\ne j$, correspondences are established by computing mutual nearest neighbors between pixel locations $\mathbf{y}_i$ and $\mathbf{y}_j$:
\begin{equation}
    \{(\mathbf{y}_i, \mathbf{y}_j) \mid \mathbf{y}_j = \text{NN}^{i,j}(\mathbf{y}_i)\ \text{and}\ \mathbf{y}_i = \text{NN}^{j,i}(\mathbf{y}_j)\},
\end{equation}
where $\mathrm{NN}^{i,j}(\mathbf{y}_i)
:= \operatorname*{arg\,min}\limits_{k\in\{1,\ldots, HW\}}
\|\mathbf{X}_i(\mathbf{y}_i)-\mathbf{X}_j(k)\|_2$  denotes the nearest neighbor of pixel $\mathbf{y}_i$ in view $j$ within 3D space. We extend this algorithm to token-level resolution by downsampling the pointmaps, and introduce a cycle consistency threshold $\tau$ to replace the exact mutual matching criterion.

\begin{table}[b]
  \centering
  \scriptsize
  \setlength{\tabcolsep}{3pt}
  \renewcommand{\arraystretch}{1.05}
  \caption{\textbf{Runtime and peak GPU memory usage across different numbers of input frames.}
  Runtime is measured in seconds, and GPU memory usage is reported in gigabytes.}
  \label{tab:runtime_mem_frames}
  \begin{tabular}{@{}c|c|ccccccccc@{}}
    \toprule
    \multicolumn{2}{c}{\textbf{Input Frames}} & 1 & 2 & 4 & 8 & 10 & 20 & 50 \\
    \midrule
    \multirow{2}{*}{\textbf{Time (s)}} & \textbf{Pointmaps}   & \textbf{0.13} & \textbf{0.16} & \textbf{0.27} & \textbf{0.52} & \textbf{0.68} & \textbf{1.73} & \textbf{5.46}  \\
                      & \textbf{Tracking}    & 1.44 & 2.95 & 6.30 & 14.10 & 18.91 & 53.10 & 304.58  \\
    \midrule
    \multirow{2}{*}{\textbf{Mem. (GB)}} & \textbf{Pointmaps}  & \textbf{2.1} & \textbf{2.4} & \textbf{3.1} & \textbf{4.5} & \textbf{5.3} & \textbf{8.8} & \textbf{19.5} \\
                       & \textbf{Tracking}   & 2.3 & 2.7 & 3.4 & 5.0 & 5.7 & 9.6 & 21.1  \\
    \bottomrule
  \end{tabular}
  \vspace{-10px}
\end{table}
\paragraph{Accuracy.}
We employ VGGT~\cite{wang2025vggt} to obtain pointmaps, as it reports superior performance compared to previous geometry prediction models, such as DUSt3R~\cite{wang2024dust3r} and MASt3R~\cite{leroy2024grounding}. As illustrated in~\cref{fig:main_analysis_a}, VGGT pointmaps achieve the correspondence precision of 83.32, significantly outperforming DINOv2 at 60.84 and Dense SIFT~\cite{5551153} at 36.43.

\paragrapht{Efficiency.}
VGGT~\cite{wang2025vggt} employs a simple feed-forward approach that build pointmaps in only 0.2 seconds. Although tracking is an alternative method for producing dense correspondences between images, dense tracking is computationally prohibitive. Moreover, tracking requires $F$ separate inferences to supervise the 3D self-attention map, where $F$ is the total number of views. We evaluate inference runtime and peak GPU memory for obtaining pointmaps and tracks from VGGT~\cite{wang2025vggt} with varying numbers of input frames. Measurements are conducted using a single NVIDIA A6000 GPU with images at $336 \times 518$ resolution. For tracking, we use 1024 query points, equivalent to the total number of query tokens in the attention map at layer $l=10$ of CAT3D~\cite{cat3d}. As shown in~\cref{tab:runtime_mem_frames}, pointmap inference is significantly faster and more memory-efficient than tracking.

\subsection{Other architectures}
\label{sec:appendix-implement_othermodel}
\paragrapht{State-of-the-art multi-view diffusion model.} We implement MVGenMaster~\cite{mvgenmaster} based on the officially released code. We use an off-the-shelf geometry model~\cite{wang2025vggt} to obtain depth maps for geometric conditions. We initialize the model with Stable Diffusion 2.1~\cite{ldm} weights, and train the model on the RealEstate10K~\cite{re10k} dataset with a batch size of 3. Other training and evaluation details are identical to those in the main model experiment. 

\paragraph{DiT-based multi-view diffusion model.} Following Matrix3D~\cite{matrix3d}, we implement a multi-view diffusion model based on a pre-trained text-to-image diffusion transformer (DiT)~\cite{li2024hunyuandit}. However, instead of using an external transformer encoder to embed conditional inputs as in Matrix3D~\cite{matrix3d}, we adopt a simpler approach by inflating the self-attention to 3D self-attention, similar to our baseline~\cite{cat3d}. Specifically, we concatenate the query, key, and value matrices of each self-attention layer, and omit the cross-attention layer and text encoder. Following Matrix3D~\cite{matrix3d}, we employ Rotary Positional Embedding (RoPE)~\cite{su2023rope} to encode each token's position and absolute sinusoidal positional encoding~\cite{visiontransformervit} to encode the viewpoint index. We use Pl\"ucker rays to represent camera poses and add the camera pose embeddings as residuals. The model is initialized with Hunyuan-DiT~\cite{li2024hunyuandit} weights and trained on the RealEstate10K~\cite{re10k} dataset with a batch size of 4. All other training and evaluation details remain identical to those of the main model.

\section{Ablation studies}
\label{sec:appendix:abl_featcost}
\begin{table}[h]
\centering
\footnotesize
\setlength{\tabcolsep}{10pt} % Increased slightly for better readability
\renewcommand{\arraystretch}{1.2}
\caption{\textbf{Ablation study for supervision target.} Evaluated on RealEstate10K~\citep{re10k} at 40k iterations.}
\label{tab:suppl_abl_featcost}

\begin{tabular}{@{} c | c c c@{}}
  \toprule
  Supervision Targets & PSNR$\uparrow$ & SSIM$\uparrow$ & LPIPS$\downarrow$ \\
  \midrule
  Feature Cost & 18.17 & 0.650 & 0.333 \\
  \textbf{Attention} & \textbf{18.42} & \textbf{0.662} & \textbf{0.323} \\
  \bottomrule
\end{tabular}
\end{table}
\vspace{-5pt}
Motivated by DIFT~\cite{tang2023dift}, which demonstrates that intermediate features from Stable Diffusion 2.1~\cite{ldm} encode correspondence, we investigate the impact of the supervision target. Specifically, we compare two approaches at layer $l=10$: (1) supervising the feature similarity map derived from the layer's output features, and (2) directly supervising the attention map. We train the models with a batch size 6, while other training and evaluation settings remain identical to the ablation experiments. \cref{tab:suppl_abl_featcost} shows that supervising the attention map yields superior performance. This confirms that geometric correspondence is most effectively regulated directly within the attention mechanism, whereas feature similarity requires computing an additional cost map extraneous to the original architecture.

\section{Qualitative results}
\label{sec:appendix-qualitative}
\label{secE}
\begin{figure*}[t]
  \centering
  \vspace{-1em}
  \includegraphics[width=\linewidth]{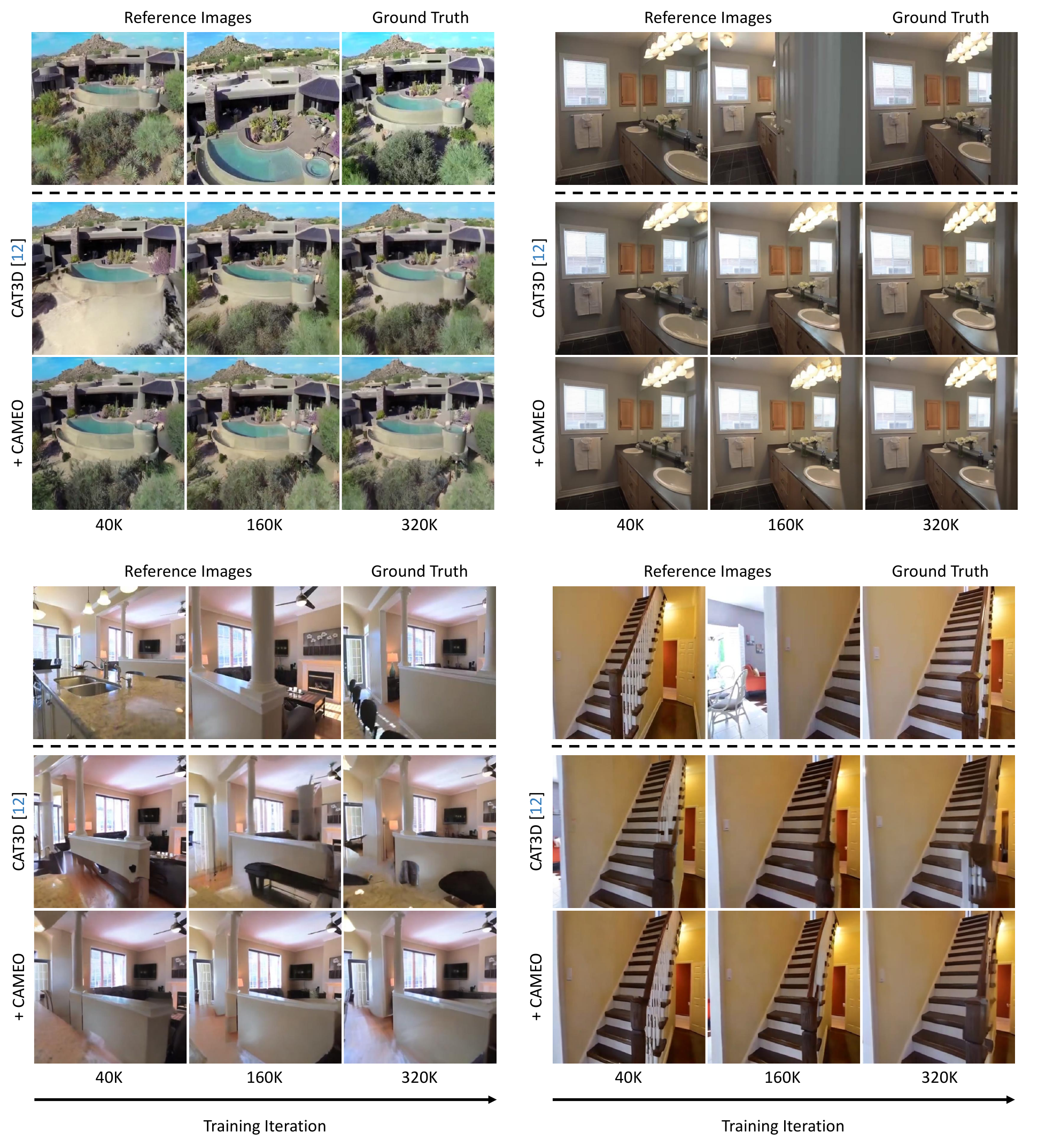}
  \caption{\textbf{Qualitative results} on RealEstate10K~\cite{re10k}.
    CAMEO improves learning efficiency while significantly enhancing geometric consistency compared to the baseline, as explicit correspondence supervision encourages faster convergence in novel view synthesis.}
  \label{fig:appendix_main_qual_re10k}
  \vspace{-10pt}
\end{figure*}

\begin{figure*}[t]
  \centering
  \vspace{-1em}
  \includegraphics[width=\linewidth]{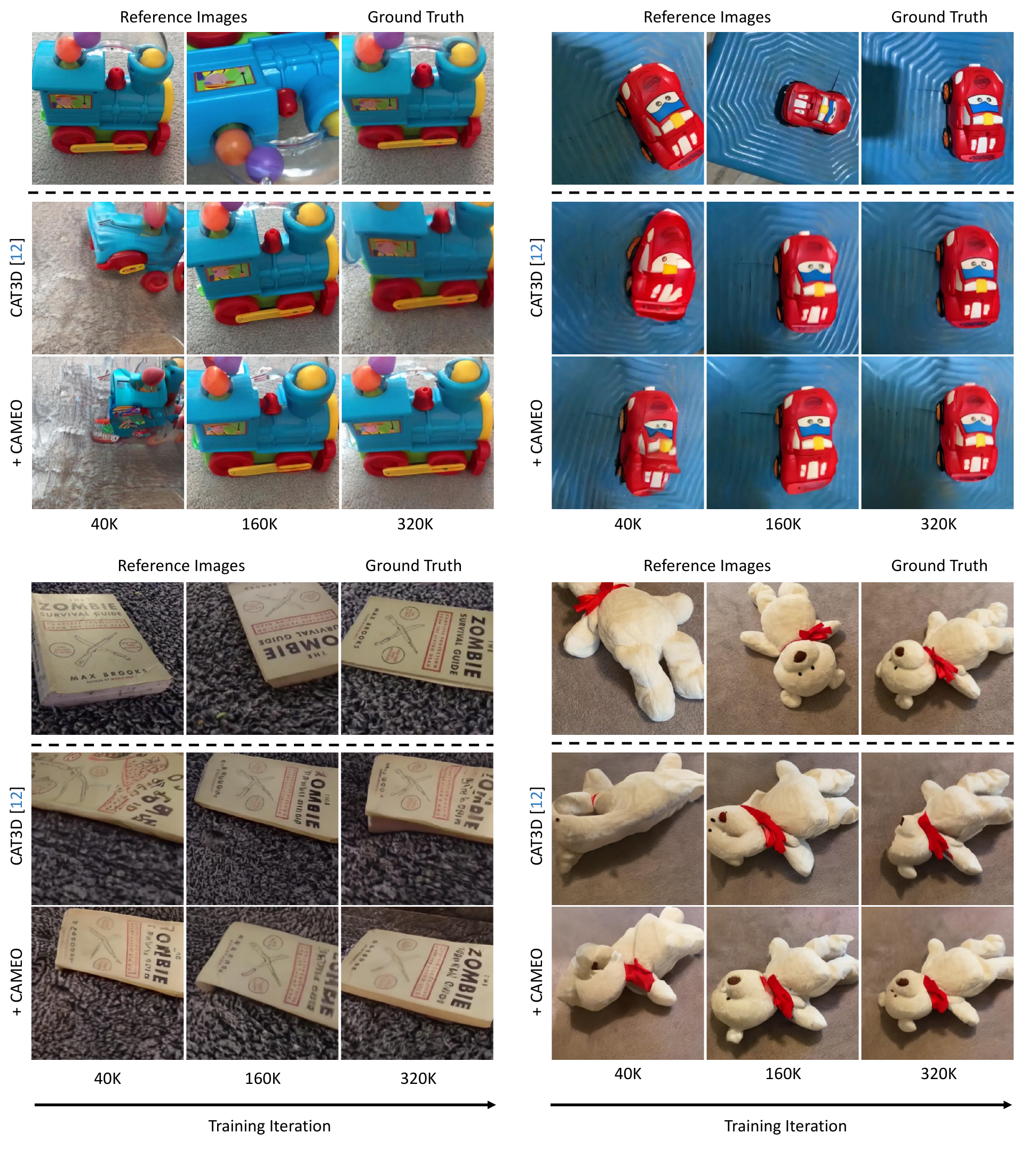}
  \caption{\textbf{Qualitative results} on CO3D~\cite{reizenstein21co3d}.
    CAMEO improves learning efficiency while significantly enhancing geometric consistency compared to the baseline, as explicit correspondence supervision encourages faster convergence in novel view synthesis.}
  \label{fig:appendix_main_co3d}
  \vspace{-10pt}
\end{figure*}

\begin{figure*}[t]
  \centering
  \vspace{-1em}
  \includegraphics[width=\linewidth]{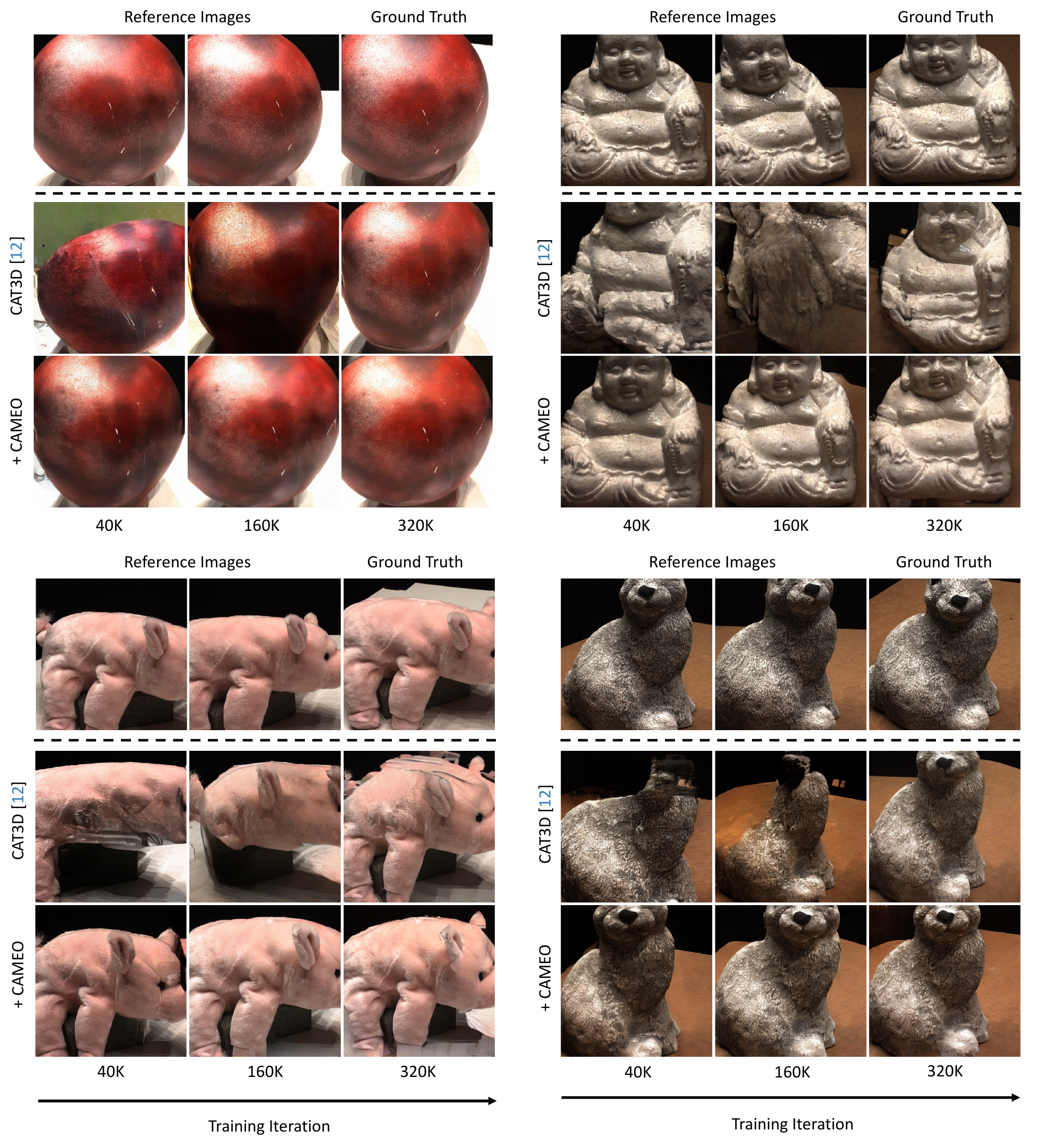}
  \caption{\textbf{Qualitative results} on DTU~\cite{jensen2014largedtu} (Out-of-domain).
    CAMEO improves learning efficiency while significantly enhancing geometric consistency compared to the baseline, as explicit correspondence supervision encourages faster convergence in novel view synthesis.}
  \label{fig:appendix_main_qual_dtu}
  \vspace{-10pt}
\end{figure*}

\begin{figure*}[t]
  \centering
  \includegraphics[width=0.88\linewidth]{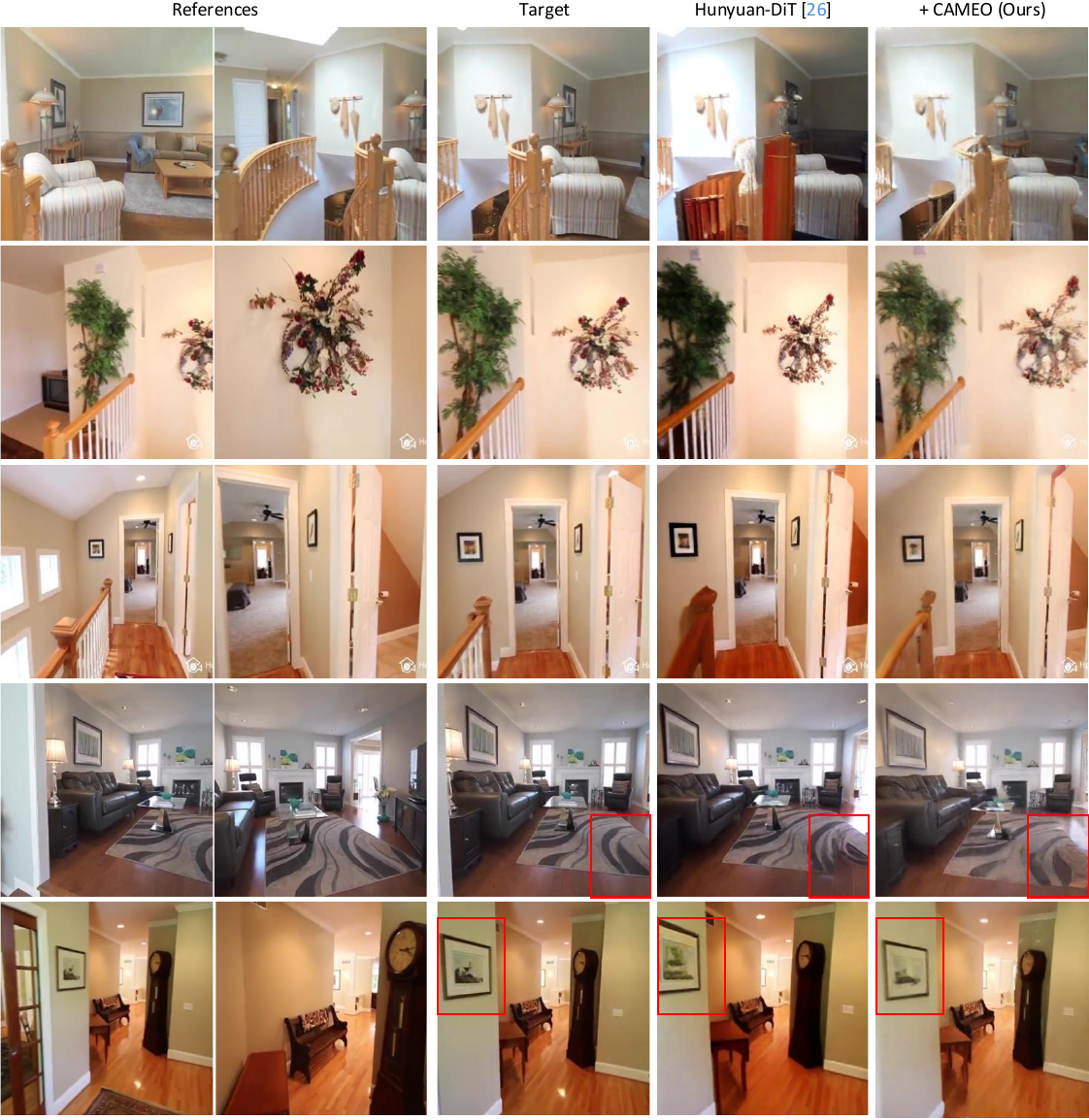}
  \caption{\textbf{Qualitative results of DiT-based model~\cite{li2024hunyuandit} on RealEstate10K~\cite{re10k}.}
    CAMEO enhances geometric consistency compared to the baseline. By incorporating explicit correspondence supervision, our method encourages the model to learn and preserve accurate structural relationships across views.}
  \label{fig:appendix_dit_qual}
\end{figure*}

\begin{figure*}[t]
  \centering
\includegraphics[width=0.88\linewidth]{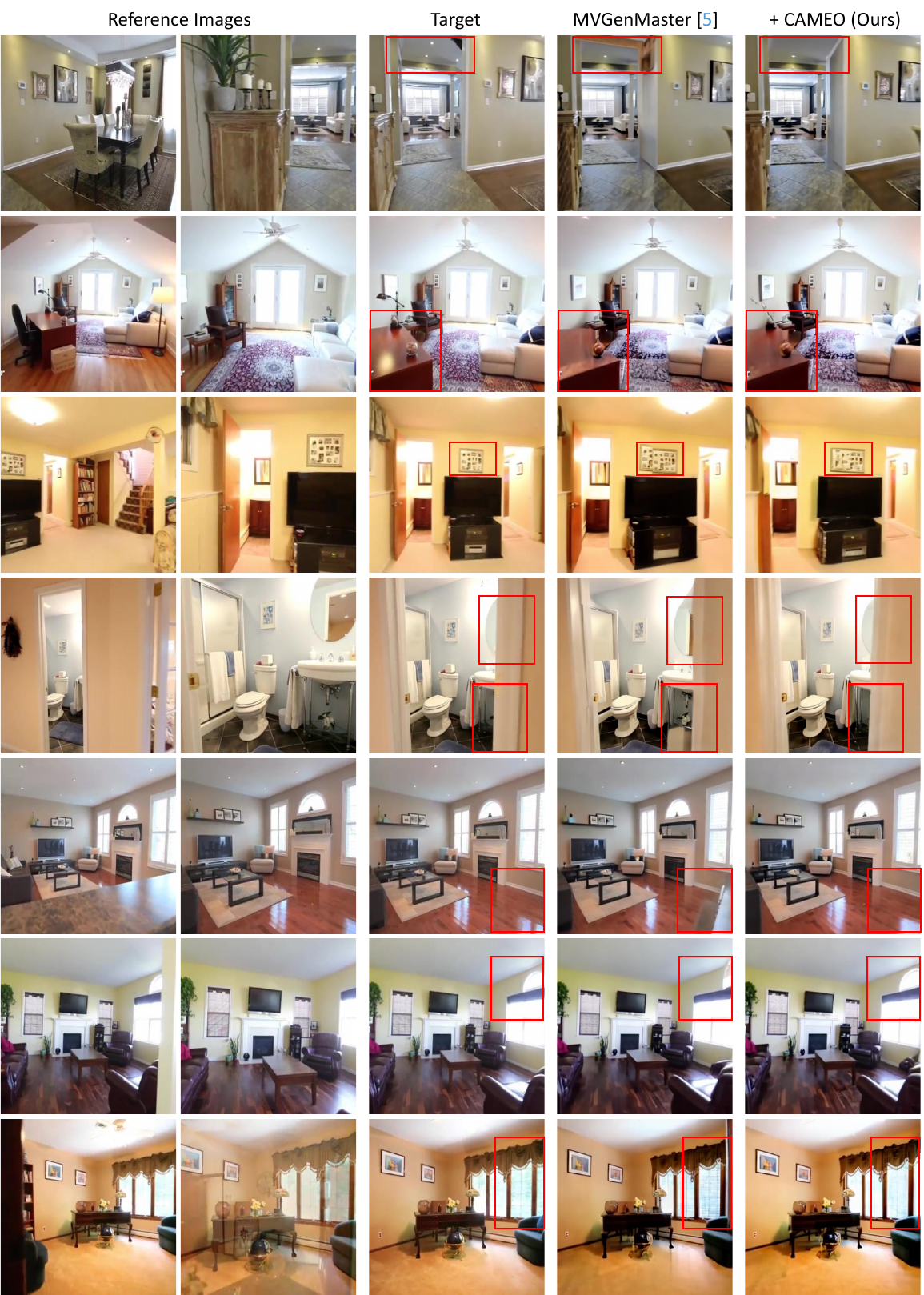}  
  \caption{\textbf{Qualitative results of MVGenMaster on RealEstate10K~\cite{re10k}.}
    CAMEO enhances geometric consistency compared to the baseline. By incorporating explicit correspondence supervision, our method encourages the model to learn and preserve accurate structural relationships across views.}
  \label{fig:appendix_mvgenmaster_qual}
\end{figure*}

We provide additional qualitative comparisons of CAMEO on both scene-level~\cite{re10k} and object-centric~\cite{reizenstein21co3d,jensen2014largedtu} settings. We present qualitative comparisons against the baseline, highlighting improved geometric consistency. \cref{fig:appendix_main_qual_re10k,fig:appendix_main_co3d,fig:appendix_main_qual_dtu} shows qualitative examples organized by training iteration.
We also provide the qualitative results of CAMEO on the state-of-the-art model~\cite{mvgenmaster} and DiT-based~\cite{li2024hunyuandit} model in~\cref{fig:appendix_dit_qual} and~\cref{fig:appendix_mvgenmaster_qual}.

\section{3D reconstruction}
\label{sec:appendix-3drecon}
% CAT3D~\cite{cat3d} proposes a pipeline that starts from sparse input views and performs 3D reconstruction~\cite{nerf}. Following this pipeline, we also optimize a 3D Gaussian Splatting (3DGS)~\cite{3dgs} using the novel views generated by our model.
Following CAT3D~\cite{3dgs}, we also perform 3D reconstruction using the novel views generated by the model. We create camera trajectories to generate novel view images and use them to optimize 3DGS~\cite{3dgs}. Specifically, we first run the multi-view diffusion models on 2-view settings, where the first and the last cameras are input views, and sample target camera trajectories between them evenly. The total number of views are 100 (2 input views and 98 generated views). We then optimize 3DGS with $\ell_1$, SSIM loss, alongside LPIPS loss following CAT3D~\cite{cat3d}, to reconstruct 3D scenes from generated novel views.

We provide the 3D reconstruction results on DTU~\cite{jensen2014largedtu} dataset in~\cref{appendix:qual-3dgs}. While CAT3D~\cite{cat3d} fails to reconstruct 3D scenes, CAMEO can faithfully reconstruct scenes through 3DGS. This demonstrates that CAMEO produces view-consistent images, leading to higher-quality 3D reconstructions than the baseline~\cite{cat3d}.

\begin{figure*}
    \centering
    \includegraphics[width=1.0\linewidth]{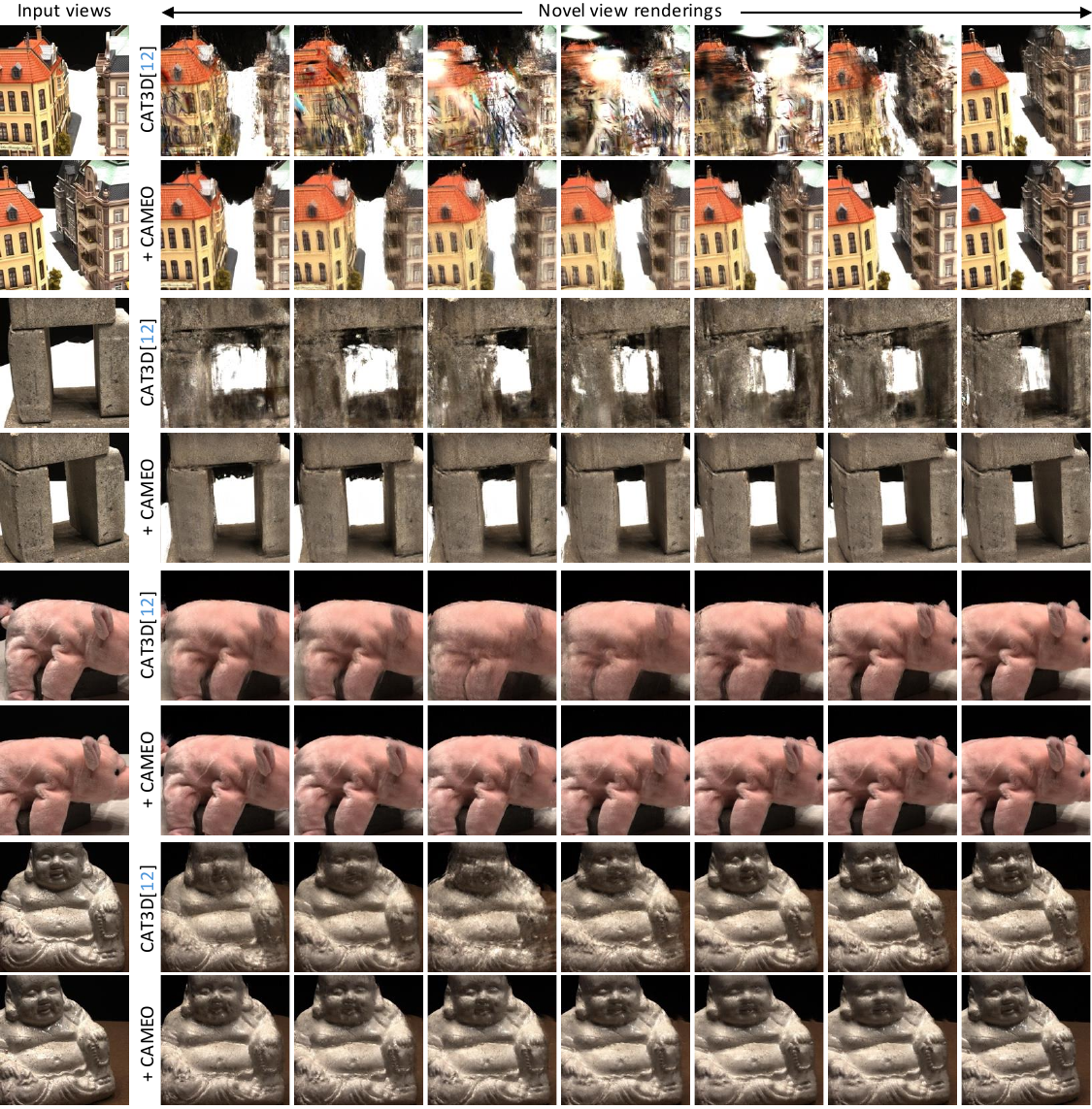}
    \caption{\textbf{3D reconstruction results.} We input 2 views and generate novel views to optimize 3DGS~\cite{3dgs}. In the 3DGS rendering results, CAMEO exhibits consistent rendered images while CAT3D~\cite{cat3d} fails.}
    \label{appendix:qual-3dgs}
\end{figure*}

\section{Limitations}
\label{sec:appendix-limitation}
\label{secF}
Our method may struggle with extreme viewpoint changes where reference and target views share minimal or no visual overlap. In such scenarios, establishing cross-view correspondence becomes inherently infeasible. Since CAMEO is designed to leverage geometric correspondences between views, its effectiveness is naturally constrained under extreme viewpoint gaps. This reflects a fundamental challenge in novel view synthesis. To address such scenarios, alternative strategies can be employed, sequentially generating intermediate views with each step conditioned on previously generated images~\cite{li2022infinitenature,Tseng_2023_CVPR}.

\section{Future work}
\label{sec:appendix-futurework}
\begin{itemize}
    \item \textbf{Beyond novel view synthesis.} Our method targets multi-view diffusion for novel view synthesis. Extending correspondence-aware supervision to video diffusion, 4D reconstruction, or other multi-modal tasks remains an open direction. 
    
    \item \textbf{Semantic correspondences.} We demonstrate that specific layers encode geometric correspondence and improve performance through geometric alignment. As an extension, semantic correspondence may be encoded in other layers, and leveraging this signal could further enhance generation quality and semantic understanding.
\end{itemize}


\begin{table*}[t]
\centering
\caption{\textbf{Performance comparison on the NAVI~\cite{navi} dataset.} We report Precision@2cm following the Probe3d~\cite{el2024probing} protocol. Best results per category are marked in \textbf{bold}. CAT3D~\cite{cat3d} and CAMEO are trained on the Co3D~\cite{reizenstein21co3d} dataset for 320k training iterations.}
\label{tab:suppl_navi}

\newcommand{\lightrule}{\arrayrulecolor{black!30}\cmidrule(lr){2-8}\arrayrulecolor{black}}

\begin{tabular}{ll c c c c c c }
\toprule
\multicolumn{2}{c}{\multirow{2}{*}{\textbf{Model}}} & \multirow{2}{*}{\textbf{Feature Grid}} & \multicolumn{5}{c}{\textbf{Precision@2cm}} \\
\cmidrule(lr){4-8}
 & & & \textbf{Overall} & $\mathbf{0-30^{\circ}}$ & $\mathbf{30-60^{\circ}}$ & $\mathbf{60-90^{\circ}}$ & $\mathbf{90-120^{\circ}}$ \\
\midrule

% VGGT
\multicolumn{2}{l}{VGGT Pointmaps~\cite{wang2025vggt}} & 518 $\times$ 518 & \textbf{83.32} & \textbf{97.41} & \textbf{92.28} & \textbf{83.12} & \textbf{63.86} \\
\multicolumn{2}{l}{Dense SIFT~\cite{5551153}}   & 128 $\times$ 128 & 36.43 & 90.02 & 53.82 & 14.51 & 8.68 \\
\midrule

% DINOv3
\multirow{2}{*}{DINOv3~\cite{simeoni2025dinov3}} & ViT-L/16 & 32 $\times$ 32 & \textbf{60.84} & \textbf{95.66} & \textbf{78.36} & \textbf{50.51} & \textbf{30.22} \\
 & ViT-B/16 & 32 $\times$ 32 & 56.68 & 94.34 & 74.91 & 44.05 & 26.18 \\

\midrule

% Naive Group 2
\multirow{11}{*}{\shortstack[l]{SD2.1~\cite{ldm}}}
 & 2  & \multirow{2}{*}{32 $\times$ 32} & 31.09 & 52.88 & 39.32 & 23.07 & 17.28 \\
 & 3  & & 28.94 & 45.68 & 36.17 & 21.97 & 18.09 \\

\lightrule

 & 4  & \multirow{2}{*}{16 $\times$ 16} & 30.75 & 57.41 & 39.66 & 22.46 & 13.55 \\
 & 5  & & 23.81 & 37.77 & 31.97 & 18.59 & 11.32 \\

\lightrule

 & 6  & 8 $\times$ 8 & 4.05 & 1.67 & 4.36 & 5.70 & 3.08 \\

\lightrule

 & 7  & \multirow{3}{*}{16 $\times$ 16} & 35.47 & 72.65 & 49.14 & 21.94 & 12.18 \\
 & 8  & & 33.04 & 71.89 & 45.55 & 19.48 & 10.33 \\
 & 9  & & 34.07 & 70.16 & 45.83 & 21.87 & 12.33 \\

\lightrule

 & 10 & \multirow{3}{*}{32 $\times$ 32} & 39.12 & 84.87 & 56.98 & 21.52 & 10.25 \\
 & 11 & & 37.95 & 80.87 & 53.71 & 21.20 & 12.42 \\
 & 12 & & \textbf{48.85} & \textbf{90.24} & \textbf{68.47} & \textbf{32.51} & \textbf{18.71} \\

 \midrule

% Naive Group 1
\multirow{11}{*}{\shortstack[l]{\textbf{CAT3D}~\cite{cat3d}}} 
 & 2  & \multirow{2}{*}{32 $\times$ 32} & 34.36 & 63.67 & 45.69 & 22.80 & 16.31 \\
 & 3  & & 33.90 & 62.19 & 44.60 & 23.31 & 16.13 \\

\lightrule

 & 4  & \multirow{2}{*}{16 $\times$ 16} & 23.28 & 40.96 & 30.43 & 17.01 & 11.17 \\
 & 5  & & 24.06 & 41.87 & 32.70 & 18.08 & 9.62 \\

\lightrule

 & 6  & 8 $\times$ 8 & 8.20 & 10.44 & 10.28 & 8.16 & 4.25 \\

\lightrule

 & 7  & \multirow{3}{*}{16 $\times$ 16} & 46.95 & 76.76 & 61.23 & 39.89 & 19.57 \\
 & 8  & & 46.07 & 78.65 & 61.05 & 36.65 & 18.94 \\
 & 9  & & 54.50 & 84.96 & 69.48 & 45.64 & 27.92 \\

\lightrule

 & 10 & \multirow{3}{*}{32 $\times$ 32} & \textbf{62.07} & \textbf{94.33} & \textbf{79.63} & \textbf{53.73} & \textbf{30.51} \\
 & 11 & & 58.61 & 92.41 & 77.70 & 48.06 & 26.77 \\
 & 12 & & 56.82 & 80.80 & 73.97 & 45.93 & 27.80 \\
\midrule

% Distill Group
\multirow{11}{*}{\shortstack[l]{\textbf{CAMEO}}}  
 & 2  & \multirow{2}{*}{32 $\times$ 32} & 35.05 & 64.84 & 46.83 & 24.28 & 15.24 \\
 & 3  & & 33.53 & 59.32 & 44.21 & 24.29 & 15.66 \\
 
\lightrule 

 & 4  & \multirow{2}{*}{16 $\times$ 16} & 23.10 & 39.91 & 30.60 & 16.85 & 10.65 \\
 & 5  & & 26.61 & 45.17 & 35.18 & 20.82 & 11.60 \\

\lightrule

 & 6  & 8 $\times$ 8 & 10.74 & 11.75 & 12.03 & 9.43 & 10.00 \\

\lightrule

 & 7  & \multirow{3}{*}{16 $\times$ 16} & 49.08 & 81.05 & 64.56 & 41.74 & 19.23 \\
 & 8  & & 49.97 & 76.29 & 62.59 & 45.38 & 23.88 \\
 & 9  & & 60.19 & 86.81 & 71.84 & 55.57 & 35.22 \\

\lightrule

 & 10 & \multirow{3}{*}{32 $\times$ 32} & \textbf{73.80} & \textbf{94.88} & \textbf{84.80} & \textbf{72.92} & \textbf{48.50} \\
 & 11 & & 63.62 & 93.23 & 79.89 & 57.12 & 33.11 \\
 & 12 & & 60.30 & 90.89 & 75.21 & 51.11 & 33.00 \\

\bottomrule
\end{tabular}
\end{table*}
\begin{table*}[t]
\centering
\small
\caption{\textbf{Correspondence precision per layer on DiT-based multi-view diffusion model~\cite{li2024hunyuandit}.} The best performing layer is in \textbf{bold}. The model is trained on the Co3D~\cite{reizenstein21co3d} dataset for 60k iterations.}
\label{tab:dit_navi}
\resizebox{\linewidth}{!}{%
\begin{tabular}{c c cccc c c c cccc}
\toprule
\multirow{2.5}{*}{\textbf{Layer}} & \multirow{2.5}{*}{\textbf{Overall}} & \multicolumn{4}{c}{\textbf{Precision@2cm}} & & \multirow{2.5}{*}{\textbf{Layer}} & \multirow{2.5}{*}{\textbf{Overall}} & \multicolumn{4}{c}{\textbf{Precision@2cm}} \\
\cmidrule(lr){3-6} \cmidrule(lr){10-13}
 & & \textbf{0--30$^\circ$} & \textbf{30--60$^\circ$} & \textbf{60--90$^\circ$} & \textbf{90--120$^\circ$} & & & & \textbf{0--30$^\circ$} & \textbf{30--60$^\circ$} & \textbf{60--90$^\circ$} & \textbf{90--120$^\circ$} \\
\midrule
0 & 11.35 & 19.27 & 13.47 & 8.57 & 7.31 & & 20 & 28.60 & 72.90 & 38.19 & 14.18 & 7.54 \\
1 & 27.90 & 72.63 & 36.92 & 13.46 & 7.37 & & 21 & 26.01 & 65.27 & 34.99 & 13.41 & 6.51 \\
2 & 27.78 & 72.68 & 37.53 & 12.64 & 7.02 & & 22 & 30.19 & 71.07 & 41.21 & 15.83 & 9.16 \\
3 & 28.05 & 71.42 & 37.53 & 13.76 & 7.51 & & 23 & 41.06 & 87.01 & 56.14 & 26.42 & 12.22 \\
4 & 32.80 & 78.74 & 45.02 & 16.11 & 10.03 & & 24 & 27.31 & 58.13 & 37.45 & 16.50 & 9.06 \\
\addlinespace
5 & 30.19 & 73.00 & 40.27 & 15.92 & 9.16 & & 25 & 26.83 & 67.62 & 36.38 & 13.32 & 6.77 \\
6 & 30.98 & 73.98 & 41.08 & 16.62 & 9.96 & & 26 & 43.62 & 83.37 & 56.45 & 32.05 & 17.69 \\
7 & 31.21 & 75.50 & 42.26 & 16.06 & 9.12 & & 27 & 30.24 & 74.41 & 39.71 & 16.56 & 8.53 \\
8 & 31.93 & 75.68 & 41.74 & 17.75 & 10.62 & & 28 & 24.84 & 67.21 & 31.61 & 12.25 & 6.41 \\
9 & 30.31 & 74.68 & 40.34 & 15.55 & 9.03 & & 29 & 48.14 & 90.51 & 64.27 & 34.97 & 18.28 \\
\addlinespace
10 & 35.78 & 80.02 & 47.87 & 20.74 & 12.24 & & 30 & 35.91 & 86.29 & 48.98 & 19.81 & 8.84 \\
11 & 31.90 & 78.01 & 43.12 & 16.28 & 9.10 & & 31 & 33.55 & 82.89 & 45.26 & 17.64 & 8.61 \\
12 & 30.84 & 76.69 & 41.19 & 15.89 & 8.52 & & \textbf{32} & \textbf{50.22} & \textbf{90.42} & \textbf{68.32} & \textbf{36.55} & \textbf{19.62} \\
13 & 33.96 & 73.14 & 45.01 & 20.81 & 12.49 & & 33 & 28.21 & 73.34 & 38.03 & 13.35 & 6.91 \\
14 & 28.05 & 64.74 & 37.32 & 14.80 & 8.87 & & 34 & 41.41 & 85.19 & 56.78 & 26.22 & 14.06 \\
\addlinespace
15 & 23.99 & 55.96 & 33.05 & 12.66 & 7.09 & & 35 & 31.79 & 74.43 & 42.44 & 17.53 & 10.13 \\
16 & 23.53 & 49.95 & 31.75 & 15.05 & 7.58 & & 36 & 34.93 & 76.04 & 48.45 & 20.60 & 10.53 \\
17 & 26.96 & 61.90 & 34.44 & 16.67 & 9.21 & & 37 & 29.22 & 72.24 & 37.82 & 15.89 & 8.91 \\
18 & 26.09 & 64.89 & 34.61 & 13.30 & 7.70 & & 38 & 30.10 & 73.30 & 41.63 & 15.26 & 7.64 \\
19 & 26.92 & 67.35 & 35.95 & 13.30 & 7.87 & & 39 & 25.94 & 69.89 & 34.34 & 11.83 & 6.25 \\
\bottomrule
\end{tabular}%
}
\end{table*}
\clearpage
{
    \small
    \enlargethispage{18cm}
    \raggedbottom
\bibliographystyle{ieeenat_fullname}
    \bibliography{main}
}
\end{document}